\documentclass[journal]{IEEEtran}
\usepackage{times}

\usepackage[numbers]{natbib}
\usepackage{multicol}
\usepackage[hyphens]{url}
\usepackage[bookmarks=true]{hyperref}
\usepackage{graphicx}
\usepackage{amsmath}
\usepackage{amssymb}
\usepackage{booktabs}
\usepackage{xcolor}
\usepackage{algorithm}
\usepackage{algorithmic}
\usepackage{multirow}
\usepackage{subcaption}
\usepackage{listings}
\usepackage{longtable}
\usepackage{array}
\usepackage{pdflscape}

\newif\ifshowcomments
\showcommentsfalse

\newcommand{\makecomment}[3]{
  \expandafter\newcommand\csname #1\endcsname[1]{%
    \ifshowcomments\textcolor{#2}{#3: ##1}\fi}}

\makecomment{will}{blue}{WS}
\makecomment{nishanth}{cyan}{NJK}
\makecomment{sahit}{purple}{SC}
\makecomment{tlp}{orange}{TLP}
\makecomment{lpk}{orange}{LPK}
\makecomment{todo}{red}{TODO}

\newcommand{\ours}{TiPToP}

\newcommand{\pddl}[1]{{\texttt{#1}}} %

\newcommand{\scenewidthland}{3cm}
\newcommand{\idpromptsep}{\vspace{3pt}\newline}

\hypersetup{
    colorlinks=true,
    linkcolor=[RGB]{70,130,208},
    citecolor=[RGB]{70,130,208},
    urlcolor=[RGB]{70,130,208},
    pdfauthor={William Shen, Nishanth Kumar, Sahit Chintalapudi, Ryan Lindeborg, Jie Wang, Christopher Watson, Edward S. Hu, Jing Cao, Dinesh Jayaraman, Leslie Pack Kaelbling, Tomas Lozano-Perez},
    pdftitle={TiPToP: A Modular Open-Vocabulary Robot Manipulation System That Plans},
    pdfsubject={Robotics, Manipulation, TAMP},
    pdfkeywords={robotics;manipulation;TAMP;planning;perception}
}

\begin{document}

\title{TiPToP: A Modular Open-Vocabulary Robot Manipulation System That Plans}


\author{
   \IEEEauthorblockN{
       William Shen\textsuperscript{1*},
       Nishanth Kumar\textsuperscript{1*},
       Sahit Chintalapudi\textsuperscript{1},
       Ryan Lindeborg\textsuperscript{1},
       Jie Wang\textsuperscript{2},
       Christopher Watson\textsuperscript{2},\\
       Edward S. Hu\textsuperscript{2},
       Jing Cao\textsuperscript{1},
       Dinesh Jayaraman\textsuperscript{2},
       Leslie Pack Kaelbling\textsuperscript{1},
       Tom\'as Lozano-P\'erez\textsuperscript{1}
   }\\[0.4em]
   \IEEEauthorblockA{
       \textsuperscript{1}MIT CSAIL,\quad
       \textsuperscript{2}University of Pennsylvania
   }
}

\makeatletter
\let\@oldmaketitle\@maketitle
\renewcommand{\@maketitle}{\@oldmaketitle
\vspace{5pt}
\centering
\setcounter{figure}{0}
\includegraphics[width=\linewidth]{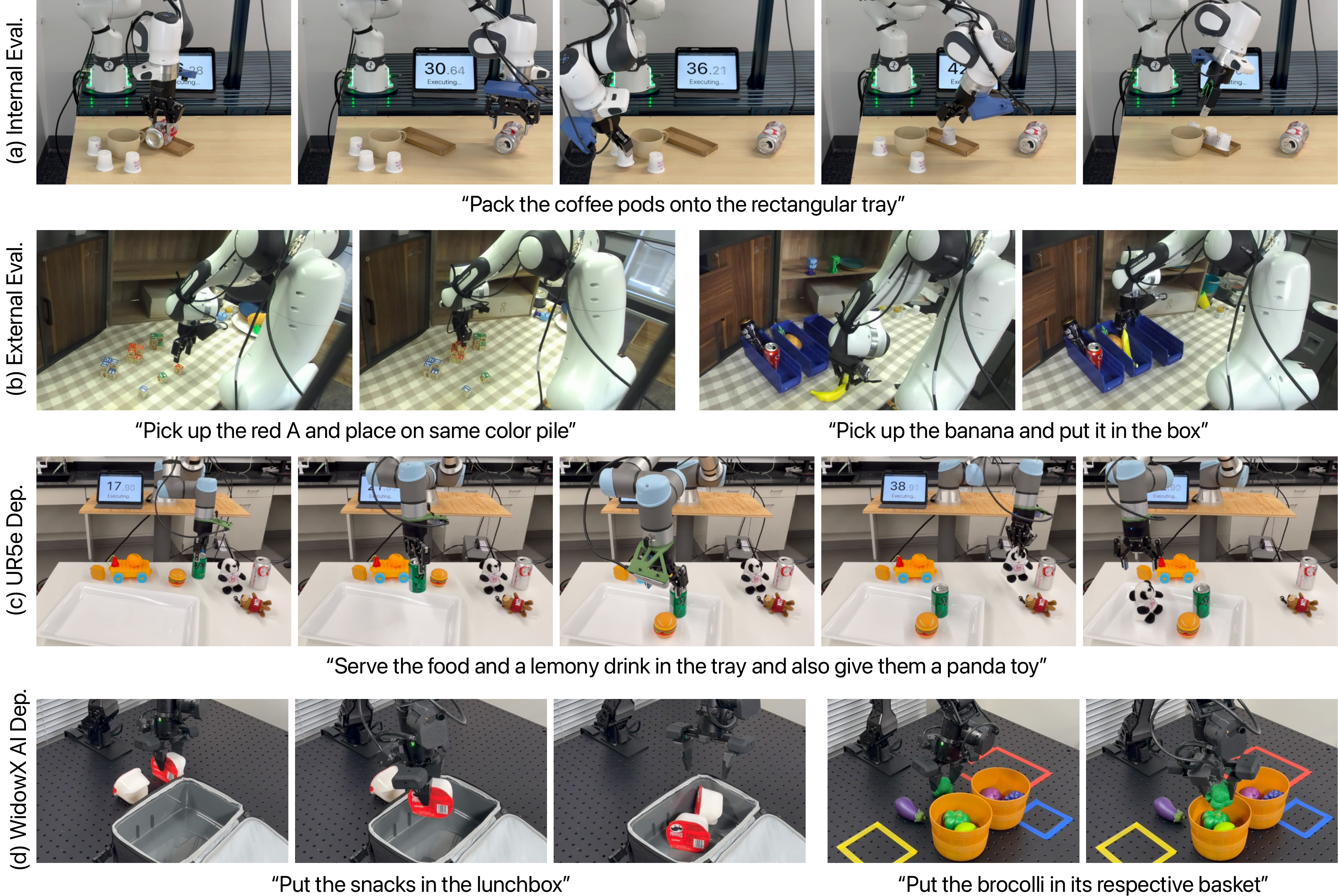}
\captionof{figure}{
\textbf{\ours{} operating over environments and embodiments.} (a) A long-horizon packing task on a DROID setup, where \ours{} first clears an obstructing Coke can. (b) Two semantic pick-and-place tasks on an external DROID setup. (c) \ours{} deployed on a UR5e and (d) on a Trossen WidowX AI.
}
\label{fig:teaser}
\vspace{-20pt}
}

\makeatother

\maketitle

\def\thefootnote{*}\footnotetext{Equal contribution. Correspondence to \texttt{\{willshen,njk\}@mit.edu}}\def\thefootnote{\arabic{footnote}}

\begin{abstract}
We present \ours{}, a modular manipulation system that integrates pretrained foundation models with a GPU-accelerated Task and Motion Planner to solve tasks directly from RGB images and natural language. \ours{} composes perception, planning, and execution modules and requires no robot training data. It can be deployed on a standard DROID setup in under an hour and adapted to new embodiments with minimal effort.
We evaluate \ours{} against $\pi_{0.5}$-DROID, a state-of-the-art VLA fine-tuned on 350 hours of demonstrations, across two real-world DROID setups (one operated by an external team) and simulation, where \ours{} attains a higher average success rate and faster average completion time. We also evaluate on the MolmoSpaces benchmark, where \ours{} ranks first overall on pick and pick-and-place tasks among methods not trained on in-distribution data.
We further show that \ours{}'s modularity enables us to trace failures to specific components, revealing where to target improvements. We release \ours{} open-source to serve as a reproducible baseline and to enable further research on modular manipulation systems.
Project website and code: \href{https://tiptop-robot.github.io}{tiptop-robot.github.io}
\end{abstract}


\IEEEpeerreviewmaketitle

\section{Introduction}

\begin{figure*}[t]
    \centering
    \includegraphics[width=\linewidth]{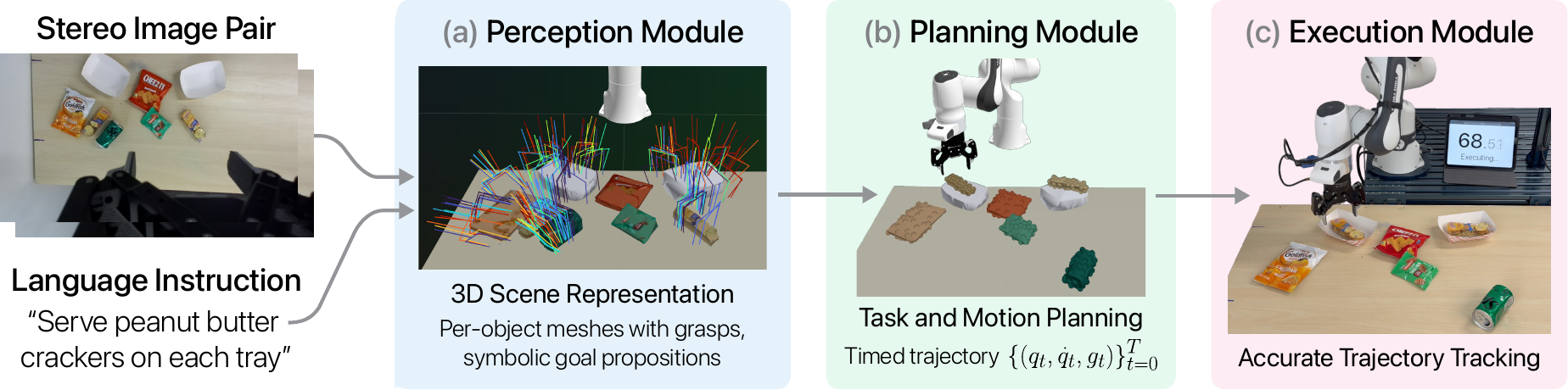}
    \caption{\textbf{\ours{} System Overview.} From a stereo RGB pair and language instruction \(\mathcal{L}\), \ours{} outputs joint trajectories with gripper commands. (a) Perception builds an object-centric 3D scene representation via depth estimation, grasp prediction, object detection, and segmentation. (b) Planning runs GPU-parallelized TAMP (cuTAMP)~\citep{shen2025cutamp} to find feasible manipulation plans. (c) Execution tracks the plan with a joint impedance controller.}
    \label{fig:pipeline}
\end{figure*}

A longstanding goal of robotics research is to build a general-purpose manipulation system that works out-of-the-box: one that can be deployed on arbitrary robots to perform language-specified tasks on arbitrary objects, with no object, environment, or embodiment-specific tuning.
Recent progress in vision and language foundation models has expanded the capabilities available to robotics, from open-vocabulary detection, grounding, and common-sense reasoning~\citep{team2023gemini, deitke2025molmo, bai2025qwen3vltechnicalreport, openai2024gpt4ocard} to depth estimation, grasp generation, and 3D shape completion~\citep{wen2025foundationstereo, yuan2023m2t2, sam3dteam2025sam3d, zadaianchuk2026recgen}. However, converting these capabilities into reliable robot manipulation behavior remains an open problem.

We introduce \textbf{\ours{}} (\textbf{T}iPToP \textbf{i}s a \textbf{P}lanner \textbf{T}hat just works \textbf{o}n \textbf{P}ixels), a complete manipulation system that pairs the perception and world knowledge of pretrained foundation models with test-time search via Task and Motion Planning (TAMP)~\citep{garrett2021integrated}. \ours{} is built on three principles. First, it is \emph{modular}: separate \emph{perception}, \emph{planning}, and \emph{execution} modules can be improved or replaced in isolation and let us trace failures to a specific module. Second, it is \emph{compositional}: a new skill requires only symbolic predicates, an operator, and a parameterized controller, which the planner composes with existing skills automatically. Third, it is \emph{zero-shot}: built from off-the-shelf models, \ours{} transfers to new embodiments, scenes, and in-skill tasks with no training or robot data. \ours{} deploys on supported robots\footnote{The embodiment must possess a camera, gripper, URDF, and trajectory tracking controller to be supported.} in under an hour with only camera calibration needed.

We evaluate \ours{} against state-of-the-art Vision-Language-Action (VLA) models~\citep{black2025pi05,lap2026,molmoact,yu2026walloss05technicalreport} and World Action Models (WAMs)~\citep{ye2026dreamzero, psi-r2, agarwal2026cosmos} given the same image and language input.
On real hardware, an external team independently deployed and evaluated \ours{} on the DROID platform~\citep{khazatsky2024droid} against $\pi_{0.5}$-DROID~\citep{black2025pi05}, a VLA fine-tuned on 350 hours of demonstrations; \ours{} matches or exceeds its success rate at comparable or shorter completion times.
On the independent MolmoSpaces benchmark~\citep{molmospaces2026}, we evaluate all 9 pick and pick-and-place tasks (9{,}000 episodes). Despite using \emph{zero} robot training data, \ours{} outperforms every VLA and WAM trained without in-distribution (ID) data on 5 of 9 tasks, and even beats ID-trained models on \emph{Pick \& Place-NextTo}.

Tracing failures to modules, we find most stem from unstable grasps under open-loop execution, pointing toward integrating \ours{}'s planning with learned, closed-loop policies that scale with data. Its compositionality and embodiment-agnostic design make it easy to extend: we add a whiteboard-wiping skill on DROID, swap in better 3D reconstruction models, and deploy on a UR5e and a Trossen WidowX AI.

Overall, we make two contributions: (1) \ours{}, an open-source, extensible planning-based manipulation system integrating foundation models with GPU-accelerated TAMP that needs no robot training data and supports real-world deployment and simulation; and (2) an empirical study and failure analysis benchmarking \ours{} against existing approaches across real-world hardware, simulation, and an independent large-scale benchmark. We release \ours{} as a reproducible, easy-to-use baseline for future systems to build on.

\section{Related Work}

\textbf{Foundation Models for Perception.}
Recent advances in vision foundation models have enabled robots to perceive diverse objects and scenes without task-specific training data.
Stereo depth estimation models~\citep{tosi2025survey, wen2025foundationstereo, guo2024stereo} predict dense depth maps from RGB image pairs. Foundation models for grasp generation~\citep{murali2025graspgen, sundermeyer2021contact, yuan2023m2t2} predict 6-DoF grasp poses from point clouds. SAM~\citep{kirillov2023segany} and SAM-2~\citep{ravi2024sam2} provide promptable segmentation from bounding boxes or points, enabling precise object boundary delineation. Shape completion models such as SAM-3D~\citep{sam3dteam2025sam3d} and RecGen~\citep{zadaianchuk2026recgen} reconstruct full 3D object meshes and poses from partial RGB(-D) views. Vision-Language Models (VLMs)~\citep{team2023gemini, deitke2025molmo, bai2025qwen3vltechnicalreport, openai2024gpt4ocard} combine vision and language understanding to perform open-vocabulary object detection, visual reasoning, and language grounding, providing semantic scene understanding and enabling robots to interpret natural language instructions.
These models each supply a perceptual capability but do not perform manipulation. SceneComplete~\citep{agarwal2025scenecomplete} composes them into a 3D scene representation to enable manipulation.
\ours{} takes this further, integrating grasp generation and a VLM with a full TAMP system to enable complex multi-step manipulation.

\textbf{Vision-Language-Action Models.}
VLAs leverage VLM backbones trained on additional diverse robot data to enable language-conditioned control~\citep{brohan2023rt2, kim2024openvla, nvidia2024groot, team2025gemini} and demonstrate that transformer-based policies trained on large datasets (e.g., the Open X-Embodiment dataset~\citep{open_x_embodiment_rt_x_2023}) can generalize across tasks and objects.

$\pi_0$~\citep{black2025pi0} introduced a flow-matching architecture, which was subsequently extended in $\pi_{0.5}$~\citep{black2025pi05} via co-training on heterogeneous data sources to improve generalization. We compare against $\pi_{0.5}$-DROID, a variant fine-tuned on 350 hours of DROID demonstrations.
A parallel line of work scales \emph{simulator}-trained VLAs: MolmoBot~\citep{molmobot2026} trains manipulation policies entirely in simulation using the MolmoSpaces ecosystem~\citep{molmospaces2026}. 
While these approaches can be applied across embodiments, show impressive generalization over scenes and objects, and can solve challenging tasks directly from pixels, they require substantial embodiment-specific training data and are trained end-to-end, making it difficult to diagnose failures.
By contrast, \ours{} composes pretrained models with no embodiment-specific training, and its modular structure allows failures to be traced to individual components. This modularity also allows individual components (perception, planning, execution) to be debugged and improved independently without modifying the entire system.

\textbf{World Models.}
World models learn to predict future observations from internet-scale video, representing dynamics directly in pixel space.
One family of approaches first generates a video plan, then recovers low-level robot actions with a separate inverse dynamics model~\citep{du2023learninguniversalpoliciestextguided, rhoda2026dva, chen2025largevideoplanner}.
World action models (WAMs) instead jointly predict video and actions~\citep{ye2026dreamzero, psi-r2, agarwal2026cosmos}.
We compare against these models, which generate an interpretable video rollout but still require fine-tuning on robot data.

\textbf{Task and Motion Planning.}
TAMP algorithms jointly solve discrete task planning and continuous motion planning~\citep{kaelbling2011hierarchical,garrett2021integrated}, complementing the open-vocabulary perception and grounding VLMs provide.
Common approaches use sampling~\citep{garrett2018ffrob,garrett2020pddlstream} or optimization~\citep{toussaint2015logic,toussaint2018differentiable} to satisfy continuous constraints, but most require detailed object geometries given \textit{a priori}.
Closest to ours, Curtis et al.~\citep{curtis2022tamp} integrate learned perception modules with PDDLStream~\citep{garrett2020pddlstream} for long-horizon manipulation of unknown objects.
\ours{} differs in three ways: (1) we use cuTAMP~\citep{shen2025cutamp}, a GPU-parallelized optimization-based planner that is far more efficient than sampling-based PDDLStream and search-then-sample bilevel planners~\citep{srivastava2014combined, chitnis2022learning}; (2) we leverage larger foundation models trained on more data; and (3) we package \ours{} to install on any embodiment providing a camera, gripper, URDF, and trajectory-tracking controller.

\textbf{Modular Robotic Planning Systems.}
Modular systems decompose manipulation into perception, high-level planning, and low-level control.
Early symbolic planners such as STRIPS~\citep{fikes1971strips} on the Shakey robot~\citep{nilsson1984shakey} showed the power of symbolic plans but required hand-engineered world models.
Recent neurosymbolic work uses LLMs to sequence pretrained skills~\citep{ahn2022saycan, huang2022innermonologue}, generate robot programs~\citep{liang2023code, singh2023progprompt}, construct 3D value maps~\citep{huang2023voxposer}, or pair VLMs with skills~\citep{Liu_2024}, and to drive TAMP by proposing task plans, subgoals, or constraints~\citep{wang2024llm3, curtis2024proc3s, Yang2024GuidingLT, kumar2024owltamp}.
Many of these systems sequence discrete skills, and none are packaged as a complete, embodiment-agnostic system deployable on real robots. \ours{} is a fully modular pipeline that directly consumes image and natural language input, jointly plans discrete task sequences and continuous collision-free trajectories, and is packaged to install easily across a variety of embodiments.

\section{Problem Formulation and System Overview}
\label{sec:problem-setting}

\begin{figure*}[t!]
    \centering
    \includegraphics[width=\linewidth]{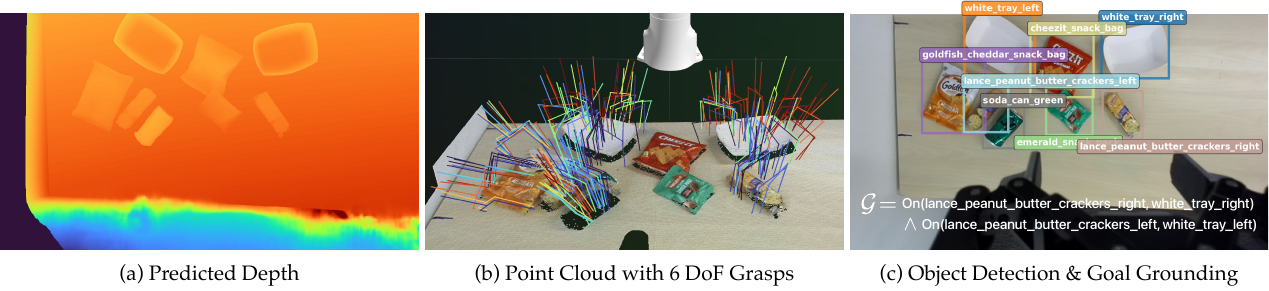}
    \caption{\textbf{Perception Module.} (a) Depth map predicted by FoundationStereo with sharp object boundaries. (b) Grasps predicted by M2T2 on the scene point cloud (colors correspond to grasp confidences). (c) Labeled object bounding boxes and symbolic goal $\mathcal{G}$ predicted by Gemini ($\texttt{On}(a, b)$ specifies that object $a$ should be placed on object or surface $b$).}
    \label{fig:perception-pipeline}
\end{figure*}

We study language-conditioned manipulation: given a natural language instruction $\mathcal{L}$ and a robot with known kinematics, produce actions that accomplish it. At each timestep $t$, a policy $\pi$ receives RGB observations $\mathbf{o}_t$ from one or more cameras and the current joint configuration $q_t$, and outputs an action $a_t$. That is,
$a_t = \pi(\mathbf{o}_t,\, q_t \mid \mathcal{L})$.
This shared input-output interface lets us directly compare two paradigms that instantiate $\pi$ in fundamentally different ways: end-to-end learned policies that map observations to actions through a single trained network, including VLAs and WAMs, and \ours{}, which composes frozen foundation models with test-time planning.

\subsection{\ours{}}

End-to-end policies instantiate $\pi$ as a single trained network that runs closed-loop at high frequency. Our representative VLA, $\pi_{0.5}$-DROID~\citep{black2025pi05}, runs at 15\,Hz: at each timestep $t$ it observes $\mathbf{o}_t = (I^{\text{wrist}}_t, I^{\text{ext}}_t)$, monocular RGB images from the wrist and external cameras, together with the current joint and gripper positions $(q_t, g_t)$, and emits chunks of 15 actions $(\dot{q}_{t:t+15}, g_{t:t+15})$, where $\dot{q}$ is a joint velocity command and $g \in \{0, 1\}$ is the binary gripper action.

\ours{} instead instantiates $\pi$ as a \emph{planner}, differing on three axes: it senses \emph{once} rather than continuously, plans a full trajectory rather than emitting short action chunks, and executes \emph{open-loop} rather than reacting to new observations\footnote{\ours{} can be made closed-loop and this is an important direction for future work (see \S\ref{sec:discussion}).}. It observes the scene \emph{once} at $t{=}0$ from a \emph{calibrated} wrist camera at a capture pose, which we assume provides a good view of the workspace. The observation $\mathbf{o}_0 = (I^{\text{left}}_0, I^{\text{right}}_0)$ is a stereo RGB image pair with known intrinsics $K$, camera-to-end-effector extrinsics $T_{\text{cam}}^{\text{ee}}$, and stereo baseline $b$. From this single observation, \ours{} produces a complete timed trajectory $a_0 = \{(q_t, \dot{q}_t, g_t)\}_{t=0}^{T}$, where $q_t$ is a joint configuration, $\dot{q}_t$ a joint velocity, and $g_t \in \{0, 1\}$ a binary gripper action. This plan is then executed open-loop with no further visual observations.

\textbf{Modular Architecture.}
\ours{} is composed of three modules (Figure~\ref{fig:pipeline}):
(1)~the \emph{perception module} (\S\ref{sec:perception}) takes $\mathbf{o}_0$ and $\mathcal{L}$ and constructs an object-centric 3D scene representation with per-object meshes, candidate grasps, and a symbolic goal $\mathcal{G}$;
(2)~the \emph{planning module} (\S\ref{sec:planning}) uses cuTAMP~\citep{shen2025cutamp} to search over plan skeletons and optimize continuous parameters (grasp poses, placement poses, collision-free trajectories) to find a feasible plan;
and (3)~the \emph{execution module} (\S\ref{sec:execution}) tracks the planned trajectory open-loop using a joint impedance controller.

\textbf{Illustrative Example.}
We illustrate \ours{} in the DROID setup (Franka FR3 with a Robotiq 2F-85 gripper and a ZED Mini stereo camera mounted on the wrist) in the following scenario: the robot is given the instruction \textit{``serve peanut butter crackers on each tray''} and the scene in Figure~\ref{fig:pipeline}.
This task requires identifying peanut butter crackers among visually similar snacks (Goldfish, Cheez-Its), requiring cultural understanding and visual knowledge to distinguish them.
Additionally, a Sprite can obstructs all grasps on the left peanut butter cracker package, complicating depth estimation due to its reflective surface and requiring the robot to move the can out of the way before grasping the crackers.

\section{Perception Module}
\label{sec:perception}

The perception module takes the initial observation \(\mathbf{o}_0\), joint configuration \(q_0\), and the language instruction \(\mathcal{L}\) as input to produce an object-centric 3D scene representation consisting of per-object meshes with candidate grasps, along with symbolic goal propositions that ground \(\mathcal{L}\) into the desired relations between objects. Two branches run in parallel: the \emph{3D Vision Branch} (\S\ref{sec:3d-vision-branch}) extracts scene geometry and grasps, while the \emph{Semantic Branch} (\S\ref{sec:semantic-branch}) identifies objects and grounds the task goal. Their outputs are then merged (\S\ref{sec:combining-outputs}).

\subsection{3D Vision Branch}
\label{sec:3d-vision-branch}

\textbf{Depth Estimation.}
We use FoundationStereo~\citep{wen2025foundationstereo}, a foundation model for stereo depth estimation, to predict a dense depth map $D$ from the stereo RGB pair \(\mathbf{o}_0 = (I^{\text{left}}_0, I^{\text{right}}_0)\) from the wrist camera, the camera intrinsics $K$, and stereo baseline $b$. $D$ is aligned to the left image $I^{\text{left}}_0$.
We found that FoundationStereo produces cleaner depth maps than the ZED camera's proprietary stereo matching, particularly on transparent, specular, and textureless surfaces (Figure~\ref{fig:perception-pipeline}a).

\textbf{Unprojecting depth to 3D.}
We unproject the depth map $D$ into a 3D point cloud using the camera intrinsics $K$, then transform the points to the world frame by composing the camera-to-end-effector extrinsics $T_{\text{cam}}^{\text{ee}}$ with the forward kinematics (FK) at the capture joint configuration $q_0$:
\vspace{-0.25em}
\[
  \mathbf{p}^{\text{world}} = T_{\text{ee}}^{\text{world}} \, T_{\text{cam}}^{\text{ee}} \, \mathbf{p}^{\text{cam}} \quad \text{where} \ T_{\text{ee}}^{\text{world}} = \text{FK}(q_0).
\vspace{-0.25em}
\]
This produces a dense point cloud of the scene in the world frame $\mathbf{p}^{\text{world}}$ (Figure~\ref{fig:perception-pipeline}b).

\textbf{Grasp Generation.}
We use M2T2~\citep{yuan2023m2t2} to predict ranked 6-DoF grasp poses from the full scene point cloud.
Object-to-grasp association is performed in \S\ref{sec:combining-outputs} using segmentation masks from the \emph{Semantic Branch} (\S\ref{sec:semantic-branch}).
M2T2 reasons over the full scene and makes predictions informed by surrounding geometry, though they are not guaranteed to be collision-free.

In our illustrative example, M2T2 generates candidate grasps on the trays, one cracker package, and the soda can (Figure~\ref{fig:perception-pipeline}b).
Note that some objects may not have predicted grasps; in such cases, we fall back to a heuristic 4-DoF grasp sampler in the \emph{planning module} (\S\ref{sec:planning}).
Having a large set of scene-level candidate grasps at this stage allows the planner to later select appropriate grasps based on task requirements and collision constraints.

We also tried GraspGen~\citep{murali2025graspgen}, but it requires segmented object point clouds and does not consider scene geometry for predicting grasps, requiring additional overhead for collision checking. We also considered AnyGrasp~\citep{fang2023anygrasp}, but its license application process complicates out-of-the-box deployment.

\subsection{Semantic Branch}
\label{sec:semantic-branch}

\textbf{Object Detection and Goal Grounding.}
We query Gemini Robotics-ER 1.5~\citep{team2025gemini}, a VLM, once to jointly extract: (1)~labels and 2D bounding boxes for objects in the scene, and (2)~a symbolic goal $\mathcal{G}$ expressed as a conjunction of \textit{predicates} (i.e., logical relations between objects) over detected objects. The system was originally developed to support just the $\texttt{On}(a, b)$ predicate, though we demonstrate defining additional predicates for new skills in \S\ref{sec:wiping}.
The VLM leverages its common-sense reasoning and cultural knowledge to ground references in the instruction to specific objects and assign task-relevant labels.

In our example with $\mathcal{L} =$ ``serve peanut butter crackers on each tray'', the VLM correctly identifies that ``peanut butter crackers'' refers to the two Lance cracker packages among other snacks (Goldfish crackers, Cheez-It crackers, nuts), and reasons that ``each tray'' requires placing one package on each, producing $\mathcal{G} = \texttt{On}(\text{crackers}_{\text{right}}, \text{tray}_{\text{right}}) \wedge \texttt{On}(\text{crackers}_{\text{left}}, \text{tray}_{\text{left}})$ (Figure~\ref{fig:perception-pipeline}c).

\textbf{Object Segmentation.}
For each detected bounding box, we use SAM-2~\citep{ravi2024sam2} to generate a pixel-level segmentation mask from $I^{\text{left}}_0$.
These masks are combined with the scene point cloud in \S\ref{sec:combining-outputs} to extract per-object geometry and assign grasps to specific objects.

\subsection{Combining Outputs}
\label{sec:combining-outputs}

We combine scene-level geometry and candidate grasps from the \textit{3D Vision Branch} with object identities and segmentation masks from the Semantic Branch into an object-centric 3D scene representation, producing per-object meshes with assigned grasps for the planning module.

\textbf{Table Detection.}
We apply RANSAC~\citep{fischler1981random} to the scene point cloud $\mathbf{p}^{\text{world}}$ to fit the dominant planar surface, which we assume to be the table.
This assumption could be relaxed by detecting multiple support surfaces (e.g., tables, floors, cabinets) via semantic segmentation or multi-plane fitting.

\begin{figure}[t]
    \centering
    \includegraphics[height=2.47cm]{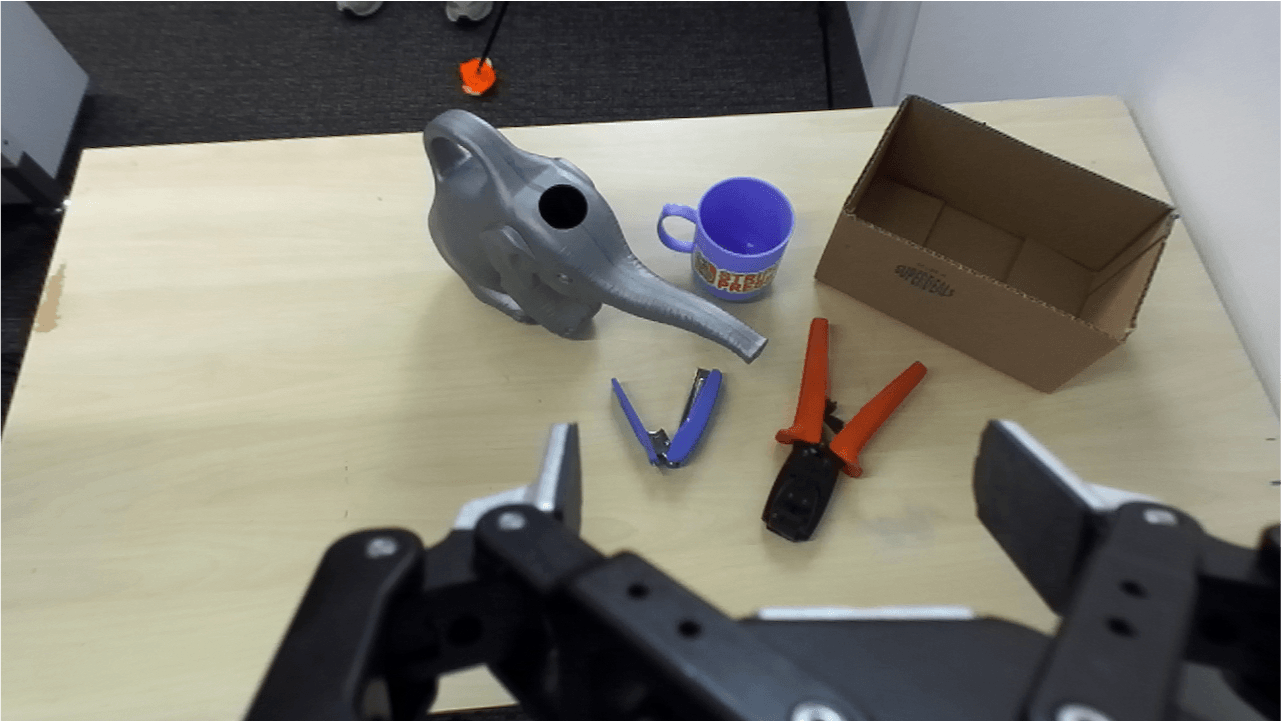}\hfill
    \includegraphics[height=2.47cm]{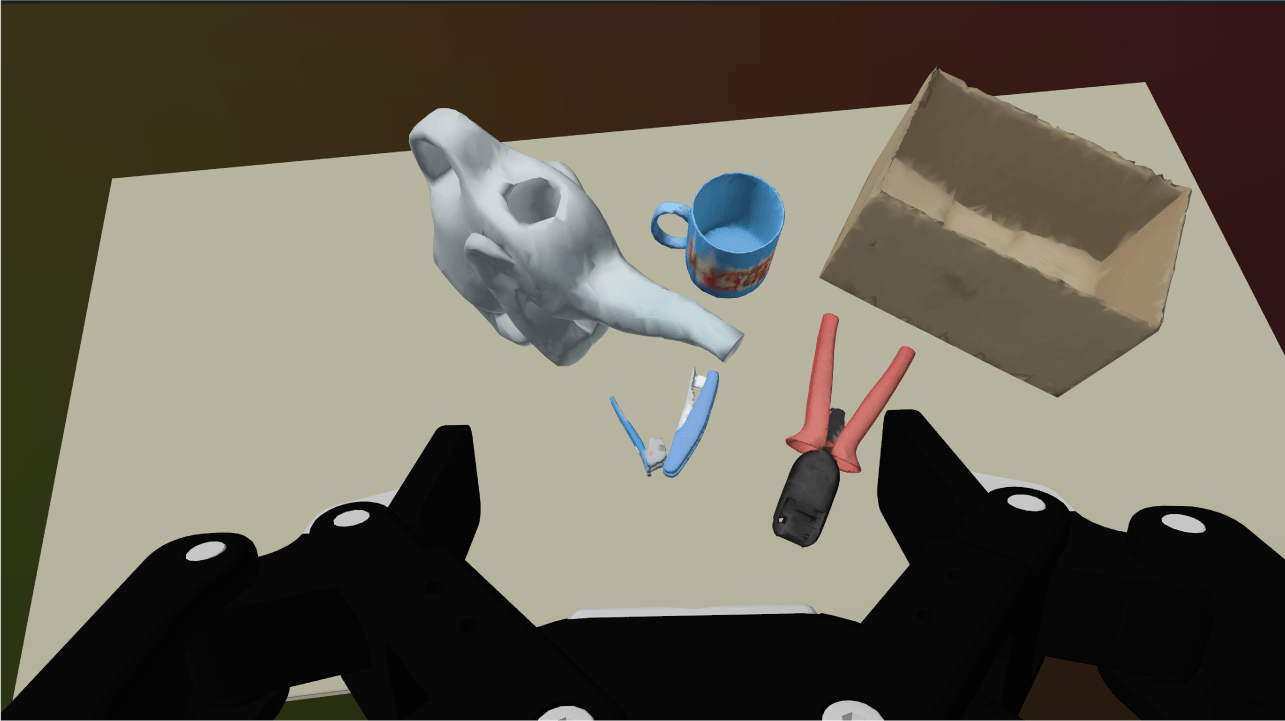}
    \caption{\textbf{Shape-Completion Mode.} From the observed scene (left), RecGen~\citep{zadaianchuk2026recgen} faithfully recovers the full geometry of each object (right), including the elephant watering can.}
    \label{fig:recgen}
\end{figure}

\textbf{Per-Object Mesh Reconstruction.}
We support two modes for reconstructing a watertight mesh for each object.
The default \emph{convex-hull} mode uses the segmentation mask of each detected object to extract the corresponding points from $\mathbf{p}^{\text{world}}$, projects them downward along the $z$-axis to the object's lowest observed point, and computes the convex hull.
We project to each object's own lowest point rather than to the table, as objects may rest on each other.
The \emph{shape-completion} mode instead feeds the RGB image $I^{\text{left}}_0$, predicted depth $D$, intrinsics $K$, and per-object segmentation masks from SAM-2 to RecGen~\citep{zadaianchuk2026recgen}, a foundation model that generates a full mesh and 6-DoF pose for each masked object (Figure~\ref{fig:recgen}).

\textbf{Design Decision: Mesh Reconstruction.}
We use the convex hull as the default because it is extremely cheap to compute and typically over-approximates the object, providing conservative geometry for collision checking. 
Although it strongly over-approximates concave objects such as bananas, we find it sufficient for the majority of tasks we evaluate. 
The shape-completion mode provides higher-fidelity meshes for complex objects at substantially greater compute cost. RecGen requires $\approx 10\,\mathrm{s}$ per object on an RTX 3090, though this trivially parallelizes over GPUs.

\textbf{Grasp-to-Object Assignment.}
Each grasp predicted by M2T2 is assigned to the nearest object by querying its contact point against a KDTree~\citep{bentley1975multidimensional} built from all object point clouds.
Grasps whose nearest object point exceeds a distance threshold are discarded, as these typically arise from point cloud noise or partial observability.

\section{Planning Module}
\label{sec:planning}

\ours{} uses cuTAMP~\citep{shen2025cutamp}, a GPU-parallelized Task and Motion Planning algorithm, to search over discrete \textit{plan skeletons} and optimize continuous parameters (grasp poses, placement poses, trajectories) to produce a full manipulation plan.
cuTAMP operates primarily over pick-and-place primitives, though it can be extended to support additional primitives such as wiping (\S\ref{sec:wiping}).
We chose cuTAMP for its fast solution times on a single GPU and its ease of installation, and made several extensions to improve its real-world deployability.

\textbf{Plan Skeleton Enumeration.}
Given the symbolic goal $\mathcal{G}$, cuTAMP uses a PDDL-style symbolic planner~\citep{mcdermott1998pddl} to enumerate candidate plan skeletons --- sequences of symbolic actions without committed continuous parameters. For example:

{\abovedisplayskip=-0.1em 
\begin{small}%
\begin{align*}
[&\pddl{MoveFree}(q_0, {?q_1}, {?\tau_1}),\; \pddl{Pick}(\pddl{cracker}, {?g}, p_0, {?q_1}), \\
&\pddl{MoveHolding}(\pddl{cracker}, {?g}, {?q_1}, {?q_2}, {?\tau_2}),\; \\
&\pddl{Place}(\pddl{cracker}, {?g}, {?p_1}, \pddl{tray}, {?q_2})]
\end{align*}
\end{small}}%
\noindent where $?g$, $?p_1$, $?q_i$, and $?\tau_i$ are unbound continuous parameters (grasp pose, placement pose, robot configurations, and trajectories, respectively).

The planner generates multiple skeletons that differ in action ordering and, crucially, may include auxiliary actions to move obstructing objects. In our example (Figure~\ref{fig:pipeline}), shorter skeletons pick and place the two cracker packages directly onto the trays, while longer skeletons additionally move the soda can out of the way before grasping the obstructed crackers.

\textbf{Particle Initialization.}
For each skeleton, cuTAMP initializes a large batch of candidate solutions, called \emph{particles}, by sampling the continuous parameters left unbound by the skeleton: grasp poses (from M2T2 predictions or a heuristic top-down grasp sampler), placement poses on target surfaces, and robot configurations via inverse kinematics. For multi-step problems, these initial samples are generally infeasible as they may violate collision, stability, or kinematic constraints.

\textbf{Particle Optimization.}
cuTAMP then ranks skeletons by a heuristic over the feasibility of their initialized particles.
For each skeleton, cuTAMP performs differentiable optimization over all particles simultaneously, refining placement poses and robot configurations to jointly satisfy collision avoidance, stable placement, and kinematic feasibility constraints. The optimization terminates once sufficient particles satisfy all constraints, moving to the next skeleton otherwise. In our example, skeletons that attempt to pick the left cracker package directly fail optimization because the soda can obstructs all feasible grasps. cuTAMP finds satisfying particles on a longer skeleton that first moves the soda can elsewhere on the table.

\textbf{Motion Planning.}
For each satisfying particle, cuTAMP invokes cuRobo~\citep{sundaralingam2023curobo}, a GPU-accelerated motion planner, to solve for the remaining trajectory parameters ($?\tau_i$) as collision-free, time-parameterized trajectories. The final output is a manipulation plan $\{(q_t, \dot{q}_t, g_t)\}_{t=0}^{T}$: joint positions, joint velocities, and gripper commands.

\textbf{Design Decision: cuTAMP Extensions.}
Deploying cuTAMP in the real world required several modifications to improve its reliability on imperfect scene reconstructions from the perception module and to extend its capabilities. Because convex-hull reconstruction over-approximates geometry, objects can appear in collision in the initial state. We exclude these false collisions from the collision cost functions so they do not block planning, increase the number of motion planning attempts, and iteratively relax collision checking thresholds. To support arbitrarily oriented surfaces, we extend cuTAMP's placement-surface cost functions to oriented bounding boxes. These and other extensions (Appendix~\ref{app:cutamp-extensions}) form a substantial part of our contribution of a working open-source system.

\section{Execution Module}
\label{sec:execution}

The execution module tracks a planned trajectory $\{(q_t, \dot{q}_t, g_t)\}_{t=0}^{T}$ on the robot. Accurately tracking trajectories is crucial, since the planner assumes consistency between the robot's joint-space execution and the resulting scene configuration. Even sub-centimeter tracking errors can cause grasps or placements to fail.

\textbf{Design Decision: Custom Joint Impedance Controller.}
Existing open-source controllers, including DROID's default Polymetis controller, could not track our timed trajectories precisely enough. We implemented our own joint-space impedance controller with tuned gains that keep the resulting end-effector tracking error within 5\,mm at high speeds, which is tight enough for our evaluation tasks. The full control law is in Appendix~\ref{app:controller-imp}.

\textbf{Design Decision: Open-loop execution.}
Our current system executes plans open-loop, without replanning from execution-time observations. We keep this first version deliberately simple to see how far a plan-once approach can go. This suffices when the scene is static and trajectories track accurately, but fails when objects move or grasps slip. Closing the loop is a natural and straightforward extension of \ours{}, and among our most impactful planned improvements (\S\ref{sec:discussion}).

\section{Experiments}
\label{sec:experiments}

Our experiments are designed to answer the following questions:
\begin{itemize}
    \item \textbf{Q1.} How well does \ours{} perform across diverse open-ended tabletop manipulation tasks, with no embodiment-specific training data?
    \item \textbf{Q2.} How efficient is \ours{} in terms of total time (planning plus execution)?
    \item \textbf{Q3.} What are the primary failure modes of \ours{}, and how are they split across the system's modules?
\end{itemize}

We evaluate \ours{} in two settings: a controlled comparison against $\pi_{0.5}$-DROID~\citep{black2025pi05}, a state-of-the-art VLA fine-tuned on 350 hours of DROID demonstrations, and on MolmoSpaces, a large-scale independent benchmark. We present each study in turn and analyze the results to answer Q1--Q3 by comparing \ours{} against existing end-to-end approaches in \S\ref{subsec:exp-discussion}.

\subsection{Controlled Comparison with \texorpdfstring{$\pi_{0.5}$}{pi-0.5}-DROID}
\label{subsec:results}

We evaluated both systems on 28 tabletop scenes across three settings: an IsaacSim simulation~\citep{NVIDIA_ISAAC_SIM} (5 tasks), the DROID setup used by \ours{}'s developers (8 tasks), and a separate DROID setup operated by an external evaluation team (15 tasks). The scenes span four categories of increasing difficulty: \emph{simple} single-step pick-and-place with no distractors; \emph{distractor} tasks that require manipulating only the relevant object amid clutter; \emph{semantic} tasks with referring expressions that demand common sense or physical reasoning about the scene (e.g., ``pick up the largest toy''); and \emph{multi-step} tasks that require sequencing several actions with physical reasoning (e.g., constrained packing, or moving an obstacle out of the way). Following an ``in-the-wild'' protocol~\citep{pi0-experiment-wild}, both systems received the same instruction and starting configuration. We report binary success rate (SR) and a finer-grained task progress (TP) defined via per-task subgoals (see Appendix~\ref{app:additional-experiment-details}).

\begin{table}[t]
\centering
\small
\setlength{\tabcolsep}{0pt}
\renewcommand{\arraystretch}{0.95}
\begin{tabular*}{\columnwidth}{@{\extracolsep{\fill}}lcccc@{}}
\toprule
& \multicolumn{2}{c}{\textbf{\ours{}}} & \multicolumn{2}{c}{\textbf{$\pi_{0.5}$-DROID}} \\
\cmidrule(lr){2-3} \cmidrule(lr){4-5}
\textbf{Scene} & SR & TP & SR & TP \\
\midrule
\multicolumn{5}{@{}l}{\emph{Simple}} \\
Cube $\to$ bowl (sim)                       & 5/10 & 72.5\%  & \textbf{8/10} & \textbf{90\%}  \\
Can $\to$ mug (sim)                         & \textbf{9/10} & \textbf{97.5\%} & 2/10 & 50\% \\
Banana $\to$ bin (sim)                      & 0/10 & 70\% & \textbf{9/10} & \textbf{97.5\%} \\
Marker $\to$ tray                           & 3/5 & 80\% & \textbf{5/5} & \textbf{100\%} \\
Crackers $\to$ tray$^\dagger$               & \textbf{5/5} & \textbf{100\%} & 3/5 & 60\%  \\
\cmidrule{2-5}
                                            & 22/40 & \textbf{84.0\%} & \textbf{27/40} & 79.5\% \\
\midrule
\multicolumn{5}{@{}l}{\emph{Distractor}} \\
Meat can $\to$ sugar box (sim)              & \textbf{5/10} & \textbf{72.5\%} & 0/10 & 5\%   \\
Coffee capsules $\to$ plate                 & \textbf{4/5} & \textbf{90\%}  & 2/5 & 58\%  \\
Turkish figs $\to$ plate                    & \textbf{3/5} & \textbf{64\%}  & 2/5 & 52\%  \\
Cashews $\to$ plate                         & 0/5 & \textbf{16\%}  & 0/5 & 12\%  \\
Red cubes $\to$ plate                       & 1/5 & 50\%  & \textbf{5/5} & \textbf{92\%} \\
Fish $\to$ box                              & \textbf{4/5} & \textbf{80\%}  & 0/5 & 10\%  \\
Crackers $\to$ tray (medium)$^\dagger$      & \textbf{5/5} & \textbf{100\%} & 3/5 & 80\%  \\
PB crackers $\to$ tray (hard)$^\dagger$     & \textbf{5/5} & \textbf{100\%} & 0/5 & 20\%  \\
\cmidrule{2-5}
                                            & \textbf{27/45} & \textbf{71.6\%} & 12/45 & 41.1\% \\
\midrule
\multicolumn{5}{@{}l}{\emph{Semantic}} \\
Toy $\to$ matching plate                    & \textbf{4/5} & \textbf{90\%}  & 1/5 & 62\%  \\
Creeper $\to$ plate                         & \textbf{3/5} & \textbf{70\%}  & 0/5 & 0\%   \\
Largest toy $\to$ plate                     & \textbf{3/5} & \textbf{70\%}  & 0/5 & 20\%  \\
Red A $\to$ color pile                      & \textbf{5/5} & \textbf{100\%} & 3/5 & 80\%  \\
Banana $\to$ box                            & \textbf{2/5} & \textbf{40\%}  & 0/5 & 30\%  \\
N block $\to$ indicated cup                 & \textbf{3/5} & \textbf{80\%}  & 2/5 & 60\%  \\
Sort blocks by color                        & \textbf{5/5} & \textbf{100\%} & 0/5 & 32\%  \\
Banana $\to$ matching plate                 & 1/5 & 20\%  & \textbf{4/5} & \textbf{90\%}  \\
\cmidrule{2-5}
                                            & \textbf{26/40} & \textbf{71.3\%} & 10/40 & 46.8\% \\
\midrule
\multicolumn{5}{@{}l}{\emph{Multi-step}} \\
Color cubes $\to$ bowl (sim)                & \textbf{9/10} & \textbf{94.6\%} & 0/10 & 24.2\% \\
AirPods $\to$ cup                           & 1/5 & 55\%  & \textbf{3/5} & \textbf{75\%}  \\
Pack pods $\to$ tray$^\dagger$              & \textbf{4/5} & \textbf{80\%}  & 1/5 & 65.7\%  \\
Pack pods $\to$ tray (obs.)$^\dagger$       & \textbf{1/5} & \textbf{67\%}  & 0/5 & 64\%  \\
Aleve bottle $\to$ tray (obs.)$^\dagger$    & \textbf{4/5} & \textbf{80\%}  & 2/5 & 70\%  \\
Three marbles $\to$ cup$^\dagger$           & \textbf{2/5} & \textbf{80\%}  & 0/5 & 6.7\%   \\
Marbles + cable$^\dagger$                   & \textbf{2/5} & \textbf{70\%}  & 0/5 & 60\%  \\
\cmidrule{2-5}
                                            & \textbf{23/40} & \textbf{75.2\%} & 6/40 & 52.2\% \\
\midrule
\textbf{Overall}                            & \textbf{98/165} & \textbf{74.6\%} & 55/165 & 52.4\% \\
\bottomrule
\end{tabular*}
\caption{
\textbf{Per-scene performance comparison over 28 scenes.} SR = Success Rate, TP = Task Progress. Best results are \textbf{bolded}. Per category and overall, SR is summed and TP averaged. $^\dagger$Evaluated by system designers; unmarked scenes evaluated by the external evaluation team.
}
\label{tab:task_performance}
\end{table}

\begin{table}[t]
\centering
\small
\renewcommand{\arraystretch}{0.95}
\begin{tabular*}{\columnwidth}{@{\extracolsep{\fill}}lccc@{}}
\toprule
& \textbf{$\pi_{0.5}$-DROID} & \multicolumn{2}{c}{\textbf{\ours{}}} \\
\cmidrule(lr){2-2} \cmidrule(lr){3-4}
\textbf{Scene} & Time (s) & Time (s) & Plan (s) \\
\midrule
\multicolumn{4}{@{}l}{\emph{Simulation}} \\
Cube $\to$ bowl    & 27.4 & \textbf{17.9} & 9.7  \\
Can $\to$ mug      & 41.0 & \textbf{18.6} & 9.2  \\
\midrule
\multicolumn{4}{@{}l}{\emph{Real-World}} \\
Crackers $\to$ tray (simple)    & 32.2 & \textbf{14.9} & 7.0 \\
Crackers $\to$ tray (medium)    & 45.2 & \textbf{14.9} & 7.3 \\
Pack pods $\to$ tray            & 53.4 & \textbf{47.0} & 20.5 \\
Aleve bottle $\to$ tray (obs.)  & 31.2 & 31.2 & 16.4 \\ 
\bottomrule
\end{tabular*}
\caption{\textbf{Completion time comparison.} `Time' reports average time-to-success over successful trials only. `Plan' reports the time required for \ours{} to run the perception and planning modules (included in `Time').}
\label{tab:time_comparison}
\end{table}

Table~\ref{tab:task_performance} reports per-scene results. Over 165 trials, \ours{} achieves a higher overall success rate (98/165 vs.\ 55/165) and task progress (74.6\% vs.\ 52.4\%) than $\pi_{0.5}$-DROID. On simple pick-and-place the two are comparable, with $\pi_{0.5}$-DROID slightly ahead on success rate (27/40 vs.\ 22/40) and \ours{} ahead on task progress. The gap widens with task complexity: \ours{} performs better on distractor, semantic, and multi-step scenes. \ours{} is also faster in total completion time (Table~\ref{tab:time_comparison}), beating $\pi_{0.5}$-DROID on five of six scenes and matching it on the sixth, often completing single-step tasks in roughly half the time.

Even when \ours{} does not fully succeed, it often completes most subgoals (e.g., 72.5\% task progress at 5/10 on cube$\to$bowl), indicating that failures tend to be isolated to a single step. $\pi_{0.5}$-DROID, by contrast, fails completely (0/5) on four of the semantic scenes.
However, $\pi_{0.5}$-DROID does better in cases where \ours{}'s convex-hull meshes misrepresent concave objects (both banana scenes), and where a slipped grasp must be retried, which \ours{}'s open-loop execution cannot do (e.g., red cubes$\to$plate: 1/5 at 50\% task progress).

\subsection{MolmoSpaces}
\label{sec:molmospaces}

MolmoSpaces~\citep{molmospaces2026} is a large-scale benchmark of diverse household tasks. We run its 9 pick and pick-and-place tasks on the DROID setup (1{,}000 episodes per task) and omit the open and close tasks, which \ours{} does not currently support. The leaderboard separates \emph{in-distribution (ID)} methods, trained on MolmoSpaces or MolmoBot simulator data, from \emph{non-ID} methods. As of the writing of this paper, \ours{} is the only non-ID method on the public leaderboard that uses zero robot or simulator demonstrations.

\begin{table*}[t]
\centering
\renewcommand{\arraystretch}{0.95}
\small
\setlength{\tabcolsep}{4pt}
\begin{tabular*}{\textwidth}{@{\extracolsep{\fill}}lcccccccccc@{}}
\toprule
& \multicolumn{2}{c}{\textbf{MolmoSpaces}} & \multicolumn{4}{c}{\textbf{MolmoBot Pick variants}} & \multicolumn{3}{c}{\textbf{MolmoBot Pick \& Place variants}} & \multicolumn{1}{c}{\textbf{Overall}} \\
\cmidrule(lr){2-3} \cmidrule(lr){4-7} \cmidrule(lr){8-10} \cmidrule(l){11-11}
\textbf{Method} & Pick & PnP & MSProc & Classic & Filam. & RandCam & PnP & PnP-NextTo & PnP-Color & {} \\
\midrule
\multicolumn{10}{@{}l}{\emph{Non-In-Distribution Methods}} \\
\textbf{\ours{} (ours)} & 68.7 & 33.2 & 67.5 & \textbf{50.0} & \textbf{48.5} & \textbf{47.8} & \textbf{29.4} & \textbf{38.0}$^{\star}$ & 31.5 & \textbf{46.1} \\
WALL-OSS-0.5 (VLA)~\citep{yu2026walloss05technicalreport}        & \textbf{73.8} & \textbf{46.0} & \textbf{78.4} & 47.9 & 47.2 & 30.1 & 13.8 & 15.9 & 18.3 & 41.3 \\
Cosmos3-Nano (WAM)~\citep{agarwal2026cosmos}            & 65.5 & 35.2 & 66.5 & 32.3 & 32.2 & 25.7 & 26.0 & 22.9 & \textbf{34.3} & 37.8 \\
Psi-R2 (WAM)~\citep{psi-r2}                & 73.1 & 41.5 & 50.2 & 23.4 & 25.8 & 27.2 & 2.6 & 2.9 & 1.5 & 27.6   \\
MolmoAct2-DROID (VLA)~\citep{molmoact}             & 48.3 & 23.3 & 43.4 & 20.5 & 21.9 & 11.1 & 15.6 & 18.5 & 19.3 & 24.7 \\
$\pi_{0.5}$-DROID (VLA)~\citep{black2025pi05}     & 36.4 & 13.6 & 18.1 & 6.4  & 7.0  & 8.0  & 12.0 & 10.3 & 10.4 & 13.6 \\
\midrule
\multicolumn{10}{@{}l}{\emph{In-Distribution Methods: fine-tuned on MolmoSpaces and MolmoBot Simulator data}} \\
MolmoBot Best (VLA)~\citep{molmobot2026}  & 92.8 & --   & 93.5 & 66.8 & 64.0 & 63.7 & 66.5 & 28.7 & 67.8 & 60.1 \\
\bottomrule
\end{tabular*}
\caption{\textbf{MolmoSpaces benchmark results} (oracle success rate \%). Best non-ID method per task in \textbf{bold}. $^{\star}$~ranks first on the full leaderboard. Integration details in Appendix~\ref{app:molmospaces-integration}.
}
\label{tab:molmospaces}
\end{table*}

Table~\ref{tab:molmospaces} reports the oracle\footnote{Oracle success rate is defined as ``success anywhere in policy execution for a time horizon, not only at the end of the episode''.} success rate against all leaderboard entrants at the time of writing. \ours{} is the strongest non-ID method overall (46.1\% SR vs.\ 41.3\% for the next-best, WALL-OSS-0.5~\citep{yu2026walloss05technicalreport}), ranking first among non-ID methods on 5 of 9 tasks. These five are the tasks with complex referring instructions and constraints that end-to-end VLAs and WAMs struggle with. On Pick \& Place-NextTo, \ours{} tops the entire leaderboard, the only task where a non-ID method beats every ID-trained model.

\subsection{Failure Analysis}
\label{subsec:failure-analysis}
\begin{figure}[t]
    \centering
    \includegraphics[width=\columnwidth]{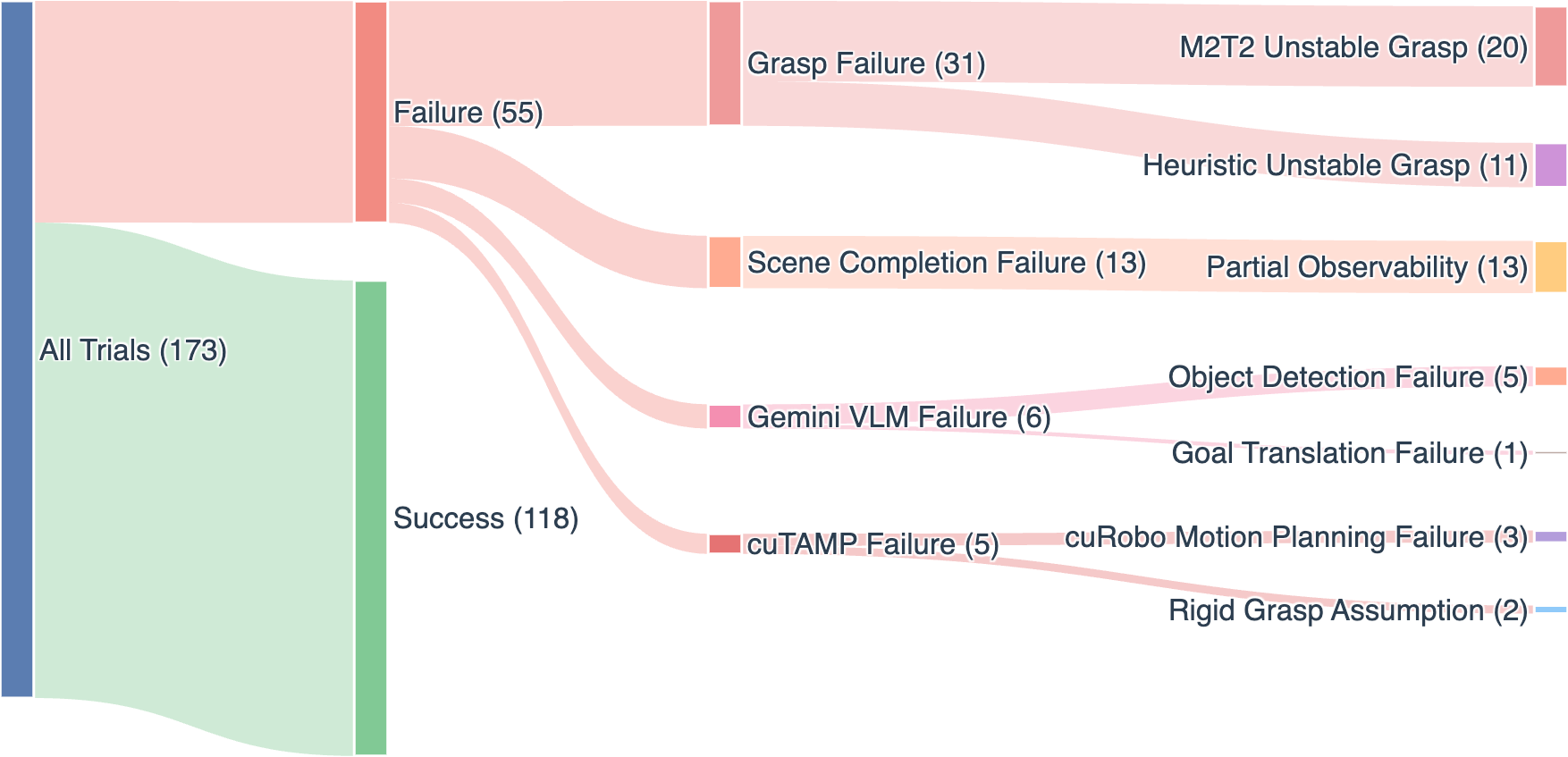}
    \caption{\textbf{Failure Analysis.} Sankey diagram showing outcomes of 173 trials. The most common failure modes are grasping failures (missed or unstable grasps), followed by scene-completion errors, VLM detection errors, then cuTAMP failures.}
    \label{fig:failure-analysis}
\end{figure}

A key advantage of \ours{}'s modular architecture is that we can trace each failure to the responsible module, something difficult to do for an end-to-end policy. Separately from the controlled comparison, we ran 173 rearrangement trials on our own DROID setup and traced the root cause of every failure to a specific module (Figure~\ref{fig:failure-analysis}).

\textbf{Grasping failures} (31/55) are the most common mode, occurring when M2T2 produces high-scoring grasps that fail in execution or when the heuristic fallback sampler is used for objects without M2T2 predictions. \textbf{Scene-completion errors} (13/55) arise from convex-hull mesh approximations that over-approximate concave objects (e.g., a banana) or under-approximate under partial observability, causing collisions during execution. \textbf{VLM errors} (6/55) occur when Gemini misdetects objects or produces incorrect bounding boxes, and \textbf{cuTAMP failures} (5/55) occur when the planner cannot find a feasible plan within its time budget. 

\subsection{Discussion}
\label{subsec:exp-discussion}

\textbf{Q1: \ours{} is competitive across diverse tasks with no robot training data.} On the MolmoSpaces benchmark, \ours{} is the strongest non-ID entrant, and in the controlled comparison it substantially outperforms $\pi_{0.5}$-DROID as task complexity grows. These gaps are largest exactly where \ours{}'s structure helps: distractor and semantic tasks benefit from explicit VLM goal grounding, which isolates the task-relevant objects amid clutter and resolves referring expressions (``largest toy,'' ``matching plate,'' ``sort by color'') that the baseline VLAs and WAMs have no mechanism for. Multi-step tasks benefit from cuTAMP's decomposition into feasible, collision-free action sequences. 
We further show this generality extends to new robots and new skills in \S\ref{sec:cross-embodiment}.

\textbf{Q2: \ours{} is efficient.} Although \ours{} spends time up front building the scene representation and planning, it is faster on nearly every timed scene in Table~\ref{tab:time_comparison}: it commits to a single time-optimal trajectory and executes it directly, whereas $\pi_{0.5}$-DROID reacts step by step, often hovering and re-attempting grasps instead of making progress. The time advantage narrows on multi-step tasks, where execution rather than planning dominates total time, although \ours{} achieves a higher success rate.

\textbf{Q3: Failures concentrate in a few modules, each addressable independently.} Our failure analysis points to two main culprits: grasping failures (over half of all failures) and convex-hull mesh approximation. Grasping failures persist because open-loop execution cannot retry a slipped or missed grasp, while the convex-hull approximation poorly represents concave shapes and is coarse for collision checking. Because \ours{} is modular, each maps to a specific component, and we suggest directions for addressing them in \S\ref{sec:discussion}.

\section{Extending to New Embodiments and Skills}
\label{sec:cross-embodiment}

\ours{}'s modular design provides several advantages. We choose to demonstrate two: the system deploys to new robot embodiments without retraining, and it accepts new manipulation skills through small, localized additions.

\textbf{Deploying to new embodiments.}
We deployed \ours{} on a UR5e arm with a wrist-mounted RealSense D435 camera (Figure~\ref{fig:teaser}c). Adapting to the new embodiment required providing the robot URDF, generating collision spheres, writing a cuRobo configuration file, and implementing camera and controller interfaces for the new hardware. The full adaptation was completed within a few hours (Appendix~\ref{app:ur5e-deployment}). We also deployed it on a Trossen WidowX AI arm with a wrist-mounted RealSense D405 camera (Figure~\ref{fig:teaser}d).

\begin{figure}[t]
    \centering
    \includegraphics[width=\columnwidth]{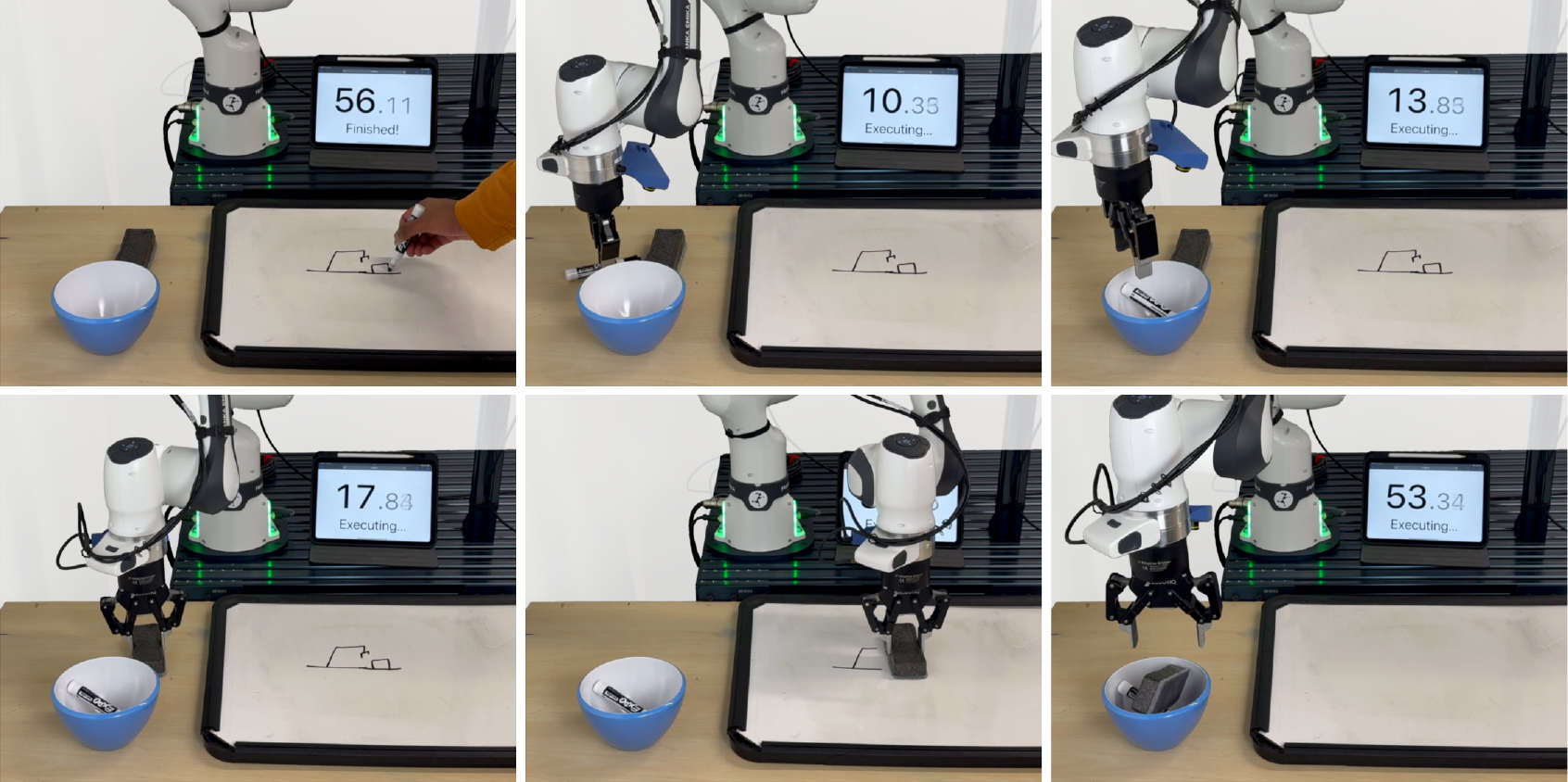}
    \caption{\textbf{Wiping.} We demonstrate that \ours{} can be straightforwardly extended to perform wiping in addition to pick-and-place. Task instruction: ``erase the whiteboard and put everything into the bowl''.}
    \label{fig:wiping-result}
\end{figure}

\textbf{Adding new skills.}
\label{sec:wiping}
We added a whiteboard-wiping primitive that erases writing from a surface with an eraser (Figure~\ref{fig:wiping-result}) through three localized changes, none of which touched the perception or execution infrastructure: (1) two new predicates (\texttt{IsEraser}, \texttt{IsCleaned}) and a goal-grounding prompt extension in the semantic branch; (2) a \texttt{Wipe} cuTAMP operator that the task planner automatically sequences after a pick; and (3) a wiping controller that localizes the region to clean and executes back-and-forth strokes over it. The entire extension took under a day (Appendix~\ref{app:wiping}).

\section{Limitations and Future Directions}
\label{sec:discussion}

\textbf{Open-loop execution.}
This is the single most impactful limitation: our failure analysis (Figure~\ref{fig:failure-analysis}) shows that grasping failures account for over half of all observed failures, many of which could be recovered from by re-attempting the grasp.
The most direct implementation to close the loop is to re-run perception and planning after each pick-and-place step, enabling recovery from failed grasps or unexpected object movement~\citep{curtis2022tamp,levihn2013foresight,pettersson2005execution,bouguerra2008monitoring}.

\textbf{Single-viewpoint perception.}
All task-relevant objects must be at least partially visible from a single wrist-camera pose.
This also limits mesh quality: with only one viewpoint, both convex hull and learned shape completion can over- or under-approximate object geometry, leading to unnecessary collisions or missed collisions during execution.
Multi-view perception, via active camera movement before planning or additional static cameras, would reduce occlusions and improve shape estimates.
Advances in depth estimation~\citep{wen2025fastfoundationstereo,tan2026masked} could further improve point cloud quality, and better grasp prediction models would also help address \ours{}'s most common observed failure mode.

\textbf{Integrating learned policies.}
Our experiments show that \ours{} and \(\pi_{0.5}\)-DROID exhibit complementary failure modes: \ours{} excels at geometric reasoning, long-horizon sequencing, and semantic grounding via its VLM goal-grounding step, but fails when grasps slip or meshes are poorly approximated; \(\pi_{0.5}\)-DROID benefits from closed-loop reactivity but struggles with multi-step structure, tight constraints, and distractor-rich scenes.
We view this complementarity as constructive. One natural way to integrate these approaches is to use end-to-end policies as reactive skills within \ours{}. This would enable robust and reactive individual skills (e.g., opening and closing articulated objects, folding, cable manipulation, contact-rich tasks) as well as constraint-aware skill chaining and long-horizon behavior.
Integrating such skills requires specifying their abstract preconditions and effects, which could be engineered or learned from data~\citep{silver2023predicate,kumar2023learning,liang2025visualpredicator,athalye2025pixelspredicates}, so the planner can reason about when to invoke them.
Another way to integrate these approaches is to call them separately for different sub-steps of long-horizon tasks (e.g., a VLA opens a drawer, then \ours{} performs multi-object pick-and-place to pack that drawer, before a VLA closes it).

\textbf{Belief-space planning.}
Extending cuTAMP to operate in belief space would enable reasoning about uncertainty in object poses, grasp outcomes, and partially observable state~\citep{kaelbling2013integrated,curtis2024partially,chintalapudi2024bilba}.
This could also enable information-gathering actions (e.g., moving the camera to observe an occluded region before planning) and more robust action selection under perceptual uncertainty.

\section{Conclusion}
We presented \ours{}, a modular planning-based manipulation system that composes pretrained vision foundation models with GPU-accelerated TAMP to solve multi-step manipulation tasks from RGB images and natural language, without any robot training data. Over 165 trials in 28 evaluation scenes in simulation and on real hardware, \ours{} matches or outperforms $\pi_{0.5}$-DROID, particularly on tasks requiring semantic grounding, distractor rejection, and multi-step sequencing. On the MolmoSpaces benchmark, \ours{} ranks first overall on pick and pick-and-place tasks among methods not trained on in-distribution data. Our system's modular architecture enables component-level failure analysis: we traced failures over 173 trials to specific modules, identifying grasping as a dominant bottleneck for the current version of the system.

A central finding of this work is that a modular system built from off-the-shelf foundation models and planning algorithms can serve as a strong manipulation system. Each component can be independently upgraded as better depth estimators, grasp predictors, VLMs, and TAMP or motion planners become available. Additionally, the complementary failure profiles of \ours{} and end-to-end policies suggest that integrating learned reactive skills within \ours{}'s framework could yield systems that combine the structured reasoning of planning with the robustness of closed-loop visuomotor control. We hope our open-source system drives further research and progress toward broadly competent and generalizable manipulation systems.

\section*{Acknowledgments}

We gratefully acknowledge support from NSF grant 2214177; from AFOSR grant FA9550-22-1-0249; from ONR MURI grants N00014-22-1-2740 and N00014-24-1-2603; from the MIT Quest for Intelligence; and from the Robotics and AI Institute.
We thank Jesse Zhang for testing TiPToP at the University of Washington. We thank Wenlong Huang for help setting up FoundationStereo to improve point cloud accuracy, as well as several helpful discussions. We thank Omar Rayyan, Maximilian Argus, Wilbert Pumacay, and Mahi Shafiullah for their encouragement and invaluable debugging support, which enabled us to integrate \ours{} with MolmoSpaces, and for adding our results to the public leaderboard. We also thank Tom Silver, Chris Agia, Joey Hejna, Karl Pertsch, Danny Driess, and Fabio Ramos for helpful discussions and feedback on earlier drafts of this work.

\subsection*{Author Contributions}

\textbf{William Shen} and \textbf{Nishanth Kumar} contributed equally to this work. William adapted and improved the core cuTAMP system to be suitable (simpler to use, faster) for our purposes. Nishanth implemented the perception interface to Gemini and SAM. Both William and Nishanth worked on integrating additional models (FoundationStereo, M2T2) into the system, packaging all components to be easily used, benchmarking system capabilities, and writing the paper. They also supported the MolmoSpaces integration and helped analyze and present results.

\textbf{Sahit Chintalapudi} implemented and packaged the control stack for the Franka Panda and FR3 robots. He also helped run quantitative experiments to investigate \ours{}'s failure modes, and helped make figures and edit the paper.

\textbf{Ryan Lindeborg} led the \ours{} integration with MolmoSpaces and gathered and analyzed the results. He also deployed \ours{} on his Trossen WidowX AI robot and provided installation and debugging feedback.

\textbf{Jie Wang} led the evaluations conducted at the University of Pennsylvania (Penn), and assisted with analysis and experimental design.

\textbf{Christopher Watson} set up \ours{} at Penn and assisted with evaluations, experimental design and analysis. 

\textbf{Edward S. Hu} assisted with \ours{} setup at Penn and contributed to experimental design and analysis. 

\textbf{Jing Cao} set up the IsaacSim simulator and ran simulation experiments comparing $\pi_{0.5}\text{-DROID}$ to \ours{}, and analyzed the results.

\textbf{Dinesh Jayaraman} advised the evaluations at the University of Pennsylvania and provided lab resources.

\textbf{Leslie Pack Kaelbling} and \textbf{Tom\'as Lozano-P\'erez} provided several helpful system implementation and task suggestions, and strongly encouraged that the code should be easy to install. They helped edit the paper, and also provided several more suggestions for improvement, the bulk of which have been left for future work.


\bibliographystyle{plainnat}
\bibliography{references}

\clearpage

\appendix

\subsection{cuTAMP Extensions}
\label{app:cutamp-extensions}

We made several extensions to cuTAMP~\citep{shen2025cutamp} to improve real-world deployability:

\textbf{M2T2 Grasp Integration.}
We support initializing grasp particles from M2T2 6-DoF grasp predictions, with collision filtering to reject grasps where the gripper would collide with the target object.

\textbf{Oriented Bounding Box Surfaces.}
We added support for oriented bounding boxes (OBBs) as placement surfaces, with cost functions that penalize object placements near surface edges, and placement samplers that account for object extents during particle initialization.

\textbf{Motion Planning Robustness.}
We increased motion planning attempts over constraint-satisfying particles since cuTAMP's collision representation (spheres) differs from cuRobo's low-level collision checks (oriented bounding boxes). When motion planning fails for path segments, we optionally relax collision checking thresholds as a fallback.

\textbf{Efficient Movable Object Collision Handling.}
We optimized collision checks between movable objects by only evaluating costs after an object's action is activated.
This allows us to handle objects that are initially in collision (e.g., due to clutter or convex hull overapproximation) by excluding them from collision penalties until moved.

\textbf{Hardware Support.}
We added robot models for the Franka FR3 with Robotiq gripper and ZED Mini camera mount, including collision sphere approximations.

\textbf{Task Planning Caching.}
Task planning becomes a bottleneck with many objects in the scene. We cache intermediate results over the task planner's tree search to reduce redundant computation.

\subsection{Controller Implementation Details}
\label{app:controller-imp}

For the DROID setup with the Franka FR3 arm, we implemented a joint impedance controller to track the planned trajectory waypoints (see \S\ref{sec:planning}) by computing joint torques at each control timestep:
\begin{equation*}
\tau = K_p \odot (q_d - q) + K_d \odot (\dot{q}_d - \dot{q}) + \tau_{\text{coriolis}} + \tau_g + M\ddot{q}_d
\end{equation*}
where $K_p$ and $K_d$ are per-joint position and velocity gains, $\tau_{\text{coriolis}}$ compensates for Coriolis forces, $\tau_g$ compensates for gravity, $M$ is the mass matrix, and $\ddot{q}_d$ is the desired acceleration estimated via filtered numerical differentiation of $\dot{q}_d$.
The term $M\ddot{q}_d$ compensates for the robot's inertia.

The gains $K_p$ and $K_d$ were tuned to improve trajectory tracking, though the controller still exhibits small deviations during execution at high speeds (typically up to 5\,mm of resulting end-effector position error).
We will open-source our controller implementation for the Franka FR3 and Panda robots upon acceptance.

For the UR5e, we instead use the \texttt{servoJ} primitive via Universal Robots' RTDE interface, with a high-gain motion phase followed by a settling phase to mitigate mechanical oscillation.

\subsection{Additional Experiment Details}
\label{app:additional-experiment-details}

Table~\ref{tab:trials} shows each evaluation scene with its language instruction and task progress metric.

\textbf{Evaluation protocol.}
$\pi_{0.5}$-DROID is a reactive policy that runs continuously until manually terminated. \ours{}, by contrast, plans once and either produces a full trajectory or explicitly fails if no valid plan is found. We use a 30--60 second planning timeout for \ours{}.

In simulation, we terminated $\pi_{0.5}$-DROID trials after 60s or upon success and reset object configurations identically across all trials for each scene.

For real-world experiments run by the external evaluators (unmarked scenes in Table~\ref{tab:trials}), $\pi_{0.5}$-DROID trials were terminated after 800 steps or upon success. They independently chose a step-based limit, which decouples evaluation from inference speed. Objects were reset to similar positions within the wrist camera's field of view using the same robot starting configuration.

For real-world experiments run by the system designers (scenes marked with $^\dagger$ in Table~\ref{tab:trials}), $\pi_{0.5}$-DROID trials were terminated after 120s or upon success. The longer timeout accommodates multi-step tasks. Since exact scene resets are not possible in the real world, we reset scenes by visually comparing against reference images, producing generally consistent configurations.

All termination limits are generous relative to typical task completion times, ensuring timeouts do not artificially limit $\pi_{0.5}$-DROID's performance. We ran all systems on an NVIDIA L4 (simulation), RTX 3080 Laptop (external evaluators), or RTX 4090 (system designers) GPU.

\textbf{Completion time.}
In Table~\ref{tab:time_comparison}, we report average completion time over successful trials only. For \ours{}, we set the \texttt{time\_dilation\_factor} in cuRobo to 0.6 in both simulation and real-world experiments. In the real-world experiments, we measure execution time using a remote iPad timer (Figure~\ref{fig:teaser}), which is automatically stopped upon robot execution for \ours{}, and manually stopped for $\pi_{0.5}$-DROID.

\begin{figure}[t]
      \centering
      \includegraphics[width=\columnwidth]{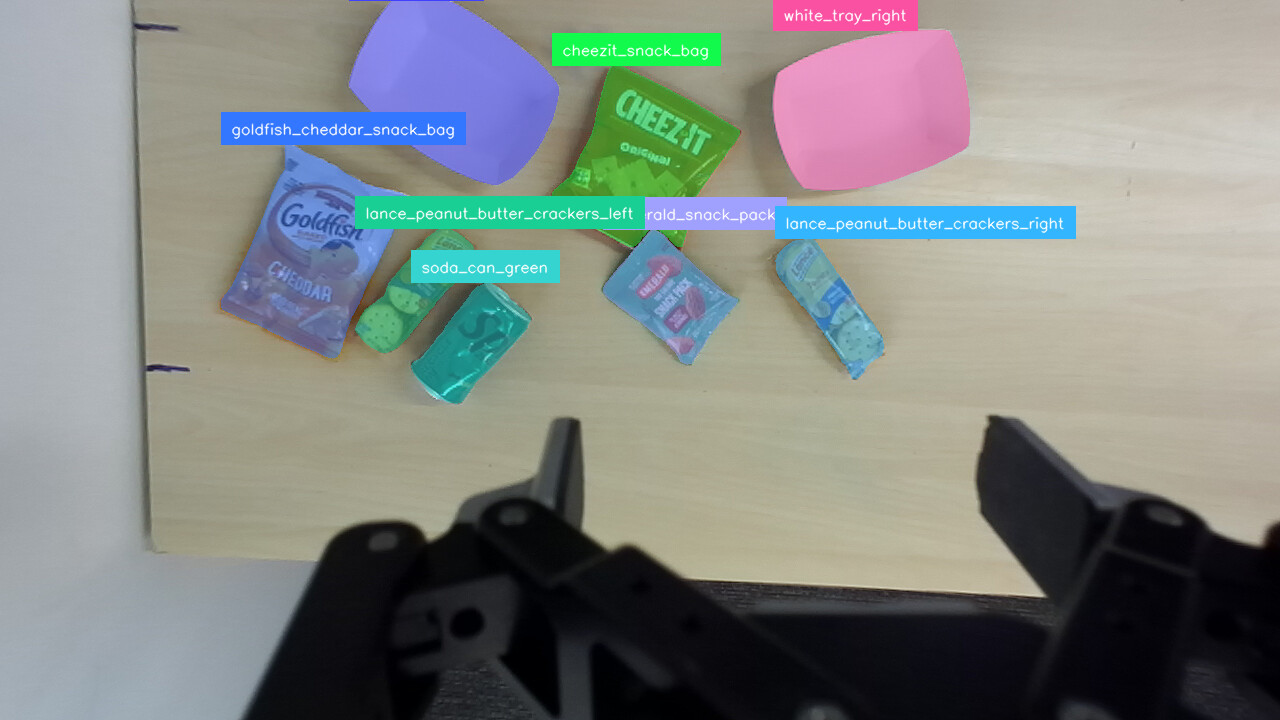}
      \caption{\textbf{Object Segmentation.} SAM-2 generates eight pixel-level segmentation masks from the bounding boxes in Fig.~\ref{fig:perception-pipeline}c.}
      \label{fig:sam-masks}
\end{figure}

\textbf{Failure Analysis.}
Our systematic failure analysis from \S\ref{subsec:failure-analysis} was performed by collecting 173 trials of \ours{} execution over a range of different tasks (different from the evaluation tasks).
For each trial, we selected a random set of objects in a random initial pose on the tabletop, and provided an appropriate natural language goal given the objects and initial configuration.
For each trial, we judged success manually and traced failures via logging and visualization.

Of the 173 trials, 52 were simple single-object pick-and-place tasks with no distractor objects (instruction: ``put the object into the container''). The remaining 121 trials all included distractor objects on the table: 50 were single-object tasks with significant clutter (instruction: ``put the smallest object into or onto the container''), 20 were single-object tasks with varied natural language goals (e.g., ``put the soft yellow object into the box'', ``put the red thing into the box'', ``put the orange item in the receptacle''), 21 were two-object pick-and-place (e.g., ``put the fruits in the orange bowl''), 20 were three-object pick-and-place (e.g., ``serve all the non-fruit food on the tray''), 5 were four-object pick-and-place (instruction: ``put all the cups with handles on the bin''), and 5 were five-object pick-and-place (instruction: ``put the caffeinated beverages and coffee pods on the box'').

\subsection{Deployment on UR5e}
\label{app:ur5e-deployment}

We deployed \ours{} on a UR5e arm with a RealSense D435 wrist camera (bottom row of Figure~\ref{fig:teaser}). Adapting \ours{} to this new embodiment required:

\begin{itemize}
    \item The robot URDF.
    \item Collision spheres for the robot geometry, automatically generated using tools like 
    \href{https://github.com/chungmin99/ballpark}{Ballpark} or \href{https://github.com/CoMMALab/foam}{Foam}.
    \item A cuRobo configuration file, following \href{https://curobo.org/tutorials/1_robot_configuration.html}{this guide} from the cuRobo developers.
    \item Code changes in cuTAMP to load the new configuration files.
    \item Code changes in \ours{} to interface with the RealSense camera (via \href{https://pypi.org/project/pyrealsense2/}{\texttt{pyrealsense2}}) and the robot controller (via Universal Robots' \href{https://sdurobotics.gitlab.io/ur_rtde/}{Real-Time Data Exchange} (RTDE) interface).
\end{itemize}

\ours{}'s codebase provides abstractions that make adding new camera types or robot controllers straightforward. Given an existing robot controller, we completed all changes in approximately 2--3 hours.

\textbf{FoundationStereo with a RealSense.} For stereo input to FoundationStereo, we used the RealSense's left and right infrared (IR) sensors. This qualitatively resulted in noisier depth estimates than the DROID setup, which uses RGB stereo pairs from the ZED Mini, particularly on transparent, specular, and reflective objects. This is expected: active IR stereo struggles with such surfaces because the projected pattern does not reflect reliably.

\textbf{Controller Implementation.} We implement a joint-space trajectory tracking controller using the Universal Robots \texttt{servoJ} primitive via the RTDE interface. The controller interpolates sparse waypoints to a 125\,Hz command stream. We use a high proportional gain (400) during motion to minimize tracking error, then reduce the gain (300) during a settling phase with dwell waypoints at the end of the trajectory to mitigate mechanical oscillation.

\subsection{Whiteboard Wiping Skill}
\label{app:wiping}

Adding the whiteboard-wiping primitive (\S\ref{sec:wiping}) required three localized changes, none of which modified the perception or execution infrastructure.

\textbf{Semantic branch.} We add two new predicates, \texttt{IsEraser} and \texttt{IsCleaned}, and extend the VLM goal-grounding prompt to translate instructions involving cleaning into conjunctions over these predicates (e.g., \texttt{IsCleaned(whiteboard)}).

\textbf{Planning.} We define a new \texttt{Wipe} cuTAMP operator with preconditions that the robot is holding an eraser and the target is a surface, and an effect that marks the surface as cleaned. The task planner automatically sequences pick then wipe to satisfy an \texttt{IsCleaned} goal. During motion solving, \texttt{Wipe} hands off to a low-level wiping skill.

\textbf{Execution.} The wiping controller calls the VLM a second time to localize the region of interest (e.g., written text) on the surface via a bounding-box query. It reprojects the bounding-box corners into world coordinates using the existing point cloud, then executes a sequence of back-and-forth strokes covering the detected region using IK-based Cartesian control.

\subsection{MolmoSpaces Integration}
\label{app:molmospaces-integration}

For the MolmoSpaces Pick and Pick \& Place-NextTo task sets, \ours{} was extended to support the predicates \texttt{Holding(?movable)} and \texttt{Near(?movable, ?reference)}. Our method can be easily extended to support new predicates for novel task types. We leave for future work adding support for the Open/Close task set variants. Other policies on the leaderboard use multiple camera views. In contrast, \ours{} only uses the wrist camera and does not need external camera views. However, \ours{} assumes that objects of interest are in the starting view of the wrist camera. Thus, we begin every trajectory by first executing a motion to a constant start pose that positions the scene within the view of the wrist camera. \ours{} relies on depth data and camera pose as inputs. Whereas in real-world experiments, this depth data is derived from stereo RGB or infrared (IR) inputs, the MolmoSpaces API allows querying for ground-truth simulation depth data directly.

\onecolumn
\small
\begin{landscape}
\setlength{\tabcolsep}{3pt}
\setlength{\LTleft}{0pt}
\setlength{\LTright}{0pt}
\begin{longtable}{@{}>{\centering}m{\scenewidthland}m{4.5cm}m{3.8cm}>{\centering}m{\scenewidthland}m{4.5cm}m{3.8cm}@{}}
\caption{%
\textbf{Evaluation scene details.} Each scene shows an image of the task, its identifier (as referenced in Table~\ref{tab:task_performance}), language prompt, and task progress metric. Scenes are grouped by category: Simple, Distractor, Semantic, and Multi-step. $^\dagger$ indicates tasks evaluated by the system designers. Unmarked scenes are evaluated by external evaluators not involved in the development of \ours{}. (sim) denotes tasks evaluated in simulation. Task progress metric numbers are reported in \%; a $+$ or $-$ sign indicates that the particular denoted amount is added or subtracted from the overall score, and no sign indicates that the number is the absolute score for achieving that particular condition.
Progress metrics may vary by the evaluator and the task. Some metrics penalize manipulating distractors while others do not.
}
\label{tab:trials} \\
\toprule
\multicolumn{1}{l}{\textbf{Scene}} & \textbf{Identifier / Language Prompt} & \textbf{Progress Metric} & \multicolumn{1}{l}{\textbf{Scene}} & \textbf{Identifier / Language Prompt} & \textbf{Progress Metric} \\
\midrule
\endfirsthead
\toprule
\multicolumn{1}{l}{\textbf{Scene}} & \textbf{Identifier / Language Prompt} & \textbf{Progress Metric} & \multicolumn{1}{l}{\textbf{Scene}} & \textbf{Identifier / Language Prompt} & \textbf{Progress Metric} \\
\midrule
\endhead
\midrule
\multicolumn{6}{r}{\emph{Continued on next page}} \\
\endfoot
\bottomrule
\endlastfoot
\multicolumn{6}{@{}l}{\emph{Simple}} \\
\includegraphics[width=\scenewidthland]{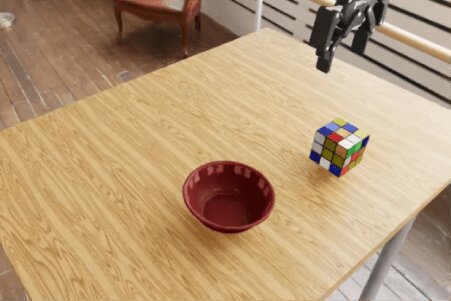} & Cube $\to$ bowl (sim)\idpromptsep ``put the cube in the bowl'' & 25\% approach cube, 50\% grasp, 75\% approach bowl with cube, 100\% place & \includegraphics[width=\scenewidthland]{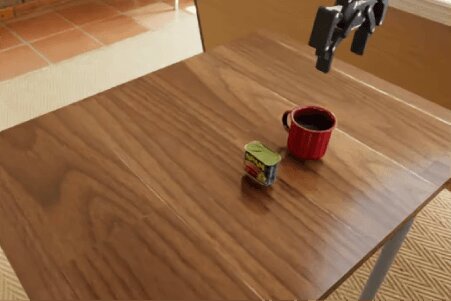} & Can $\to$ mug (sim)\idpromptsep ``put the can in the mug'' & 25\% approach can, 50\% grasp, 75\% approach mug with can, 100\% place \\
\includegraphics[width=\scenewidthland]{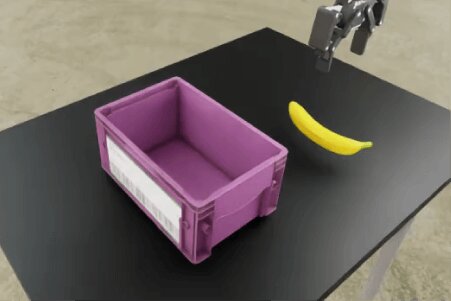} & Banana $\to$ bin (sim)\idpromptsep ``put banana in the bin'' & 25\% approach banana, 50\% grasp, 75\% approach bin with banana, 100\% place & \includegraphics[width=\scenewidthland]{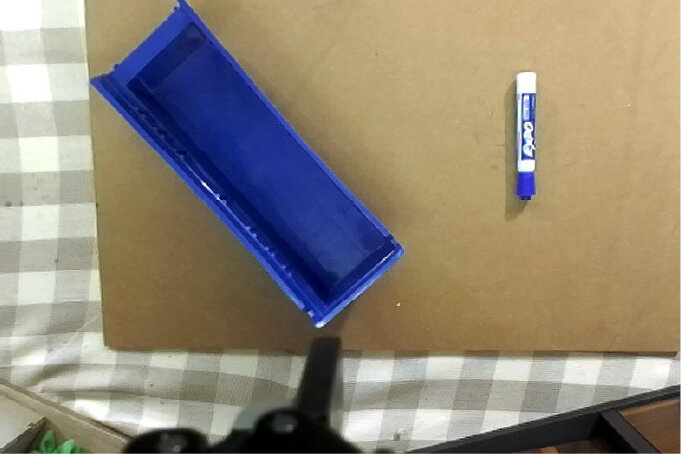} & Marker $\to$ tray\idpromptsep ``put the marker in the tray'' & +25\% touch marker, +25\% grasp, +25\% touch tray, +25\% place \\
\includegraphics[width=\scenewidthland]{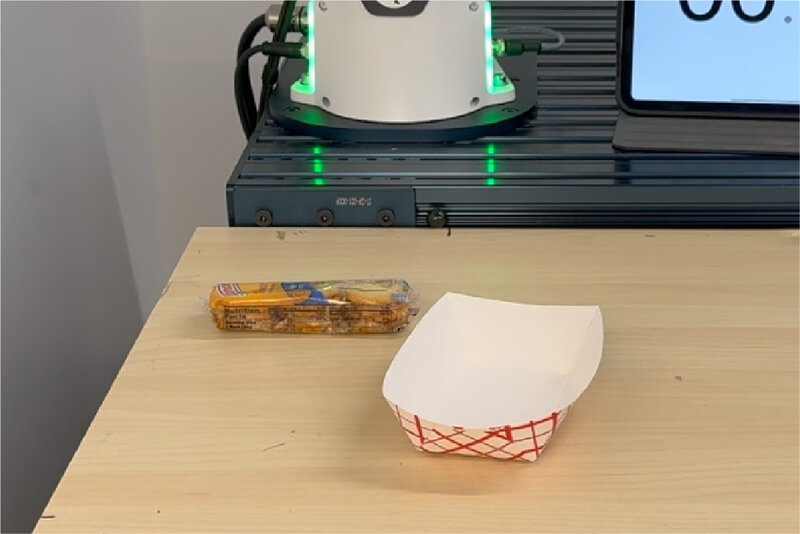} & Crackers $\to$ tray$^\dagger$\idpromptsep ``place the crackers onto the tray'' & 50\% grasp crackers, 100\% place &  &  &  \\
\midrule
\multicolumn{6}{@{}l}{\emph{Distractor}} \\
\includegraphics[width=\scenewidthland]{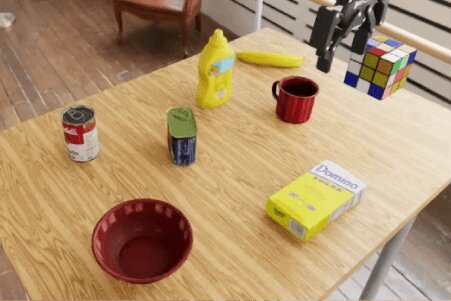} & Meat can $\to$ sugar box (sim)\idpromptsep ``put the meat can on the sugar box'' & 25\% approach meat can, 50\% grasp, 75\% approach box with meat can, 100\% place & \includegraphics[width=\scenewidthland]{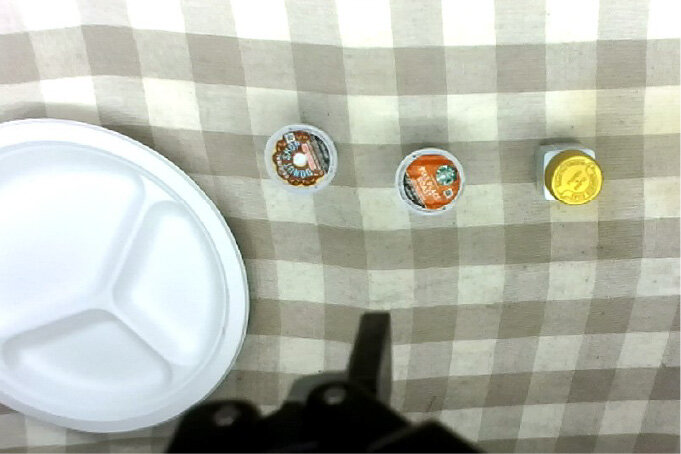} & Coffee capsules $\to$ plate\idpromptsep ``put all of the coffee capsules onto the white plate'' & +50\% per capsule placed, $-$20\% per distractor \\
\includegraphics[width=\scenewidthland]{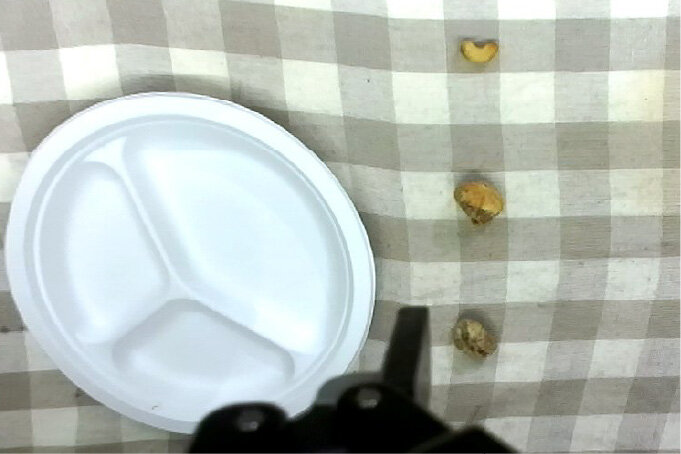} & Turkish figs $\to$ plate\idpromptsep ``put the turkish figs onto the white plate'' & +50\% per fig placed, $-$20\% per cashew & \includegraphics[width=\scenewidthland]{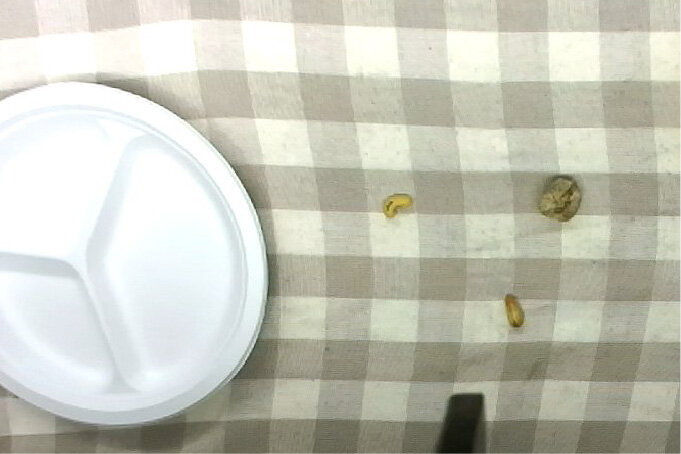} & Cashews $\to$ plate\idpromptsep ``put the roasted cashews onto the white plate'' & +50\% per cashew placed, $-$20\% per fig \\
\includegraphics[width=\scenewidthland]{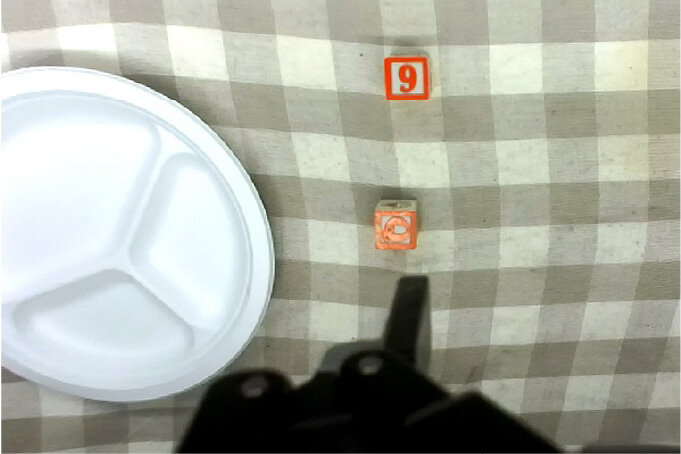} & Red cubes $\to$ plate\idpromptsep ``put the red cubes onto the white plate'' & +50\% per cube placed, $-$20\% if distractor placed & \includegraphics[width=\scenewidthland]{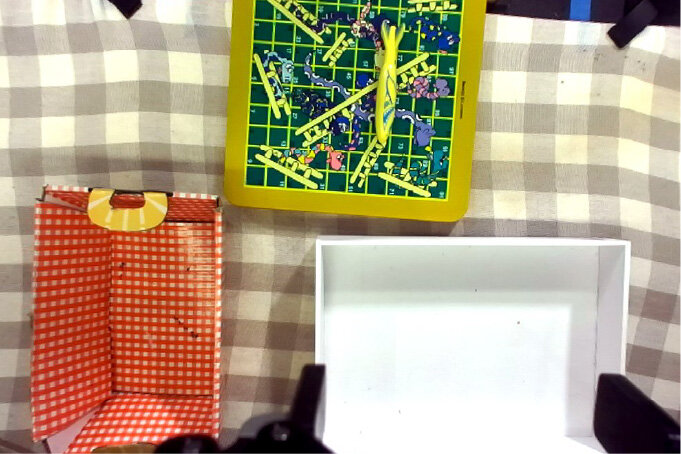} & Fish $\to$ box\idpromptsep ``place the fish into the white box'' & +50\% pick fish, +50\% place into white box \\
\includegraphics[width=\scenewidthland]{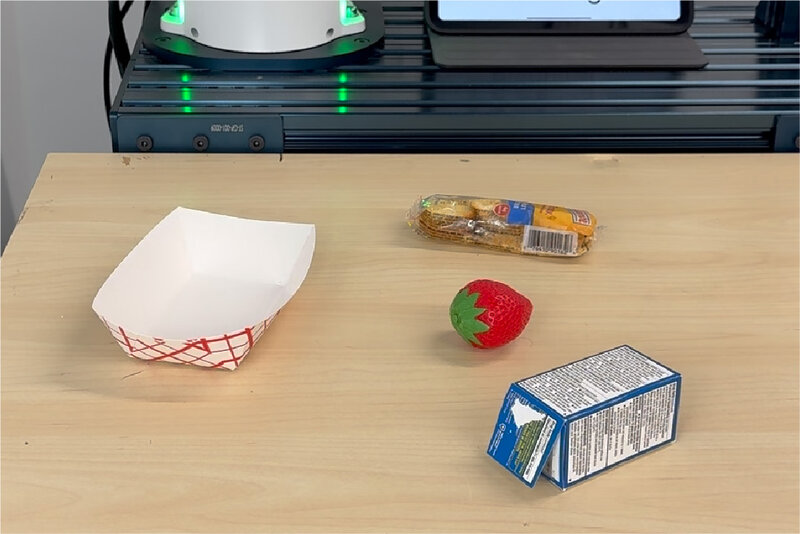} & Crackers $\to$ tray (med.)$^\dagger$\idpromptsep ``place the crackers onto the tray'' & +50\% pick crackers, +50\% place on the tray (no penalty for distractor) & \includegraphics[width=\scenewidthland]{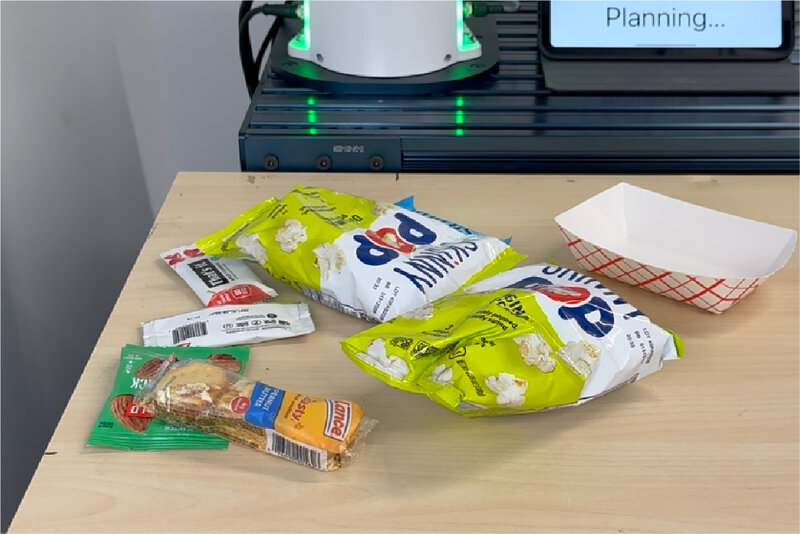} & PB crackers $\to$ tray (hard)$^\dagger$\idpromptsep ``place the peanut butter crackers onto the tray'' & +50\% pick crackers, +50\% place on the tray (no penalty for distractor) \\
\midrule
\multicolumn{6}{@{}l}{\emph{Semantic}} \\
\includegraphics[width=\scenewidthland]{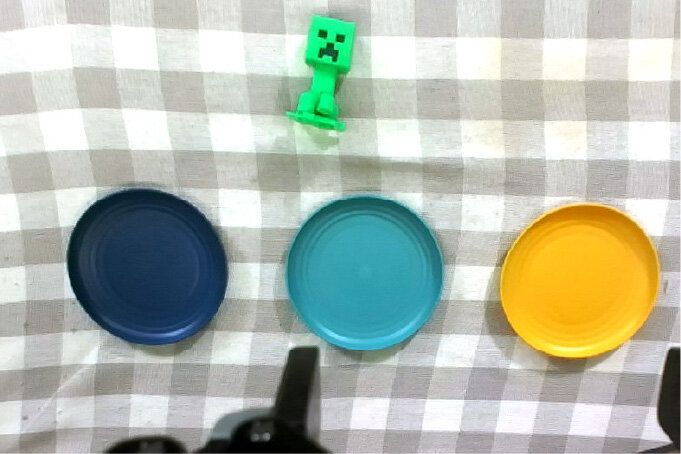} & Toy $\to$ matching plate\idpromptsep ``pick up the toy and place on the plate with similar color'' & +50\% pick toy, +50\% place on teal or +30\% place on blue & \includegraphics[width=\scenewidthland]{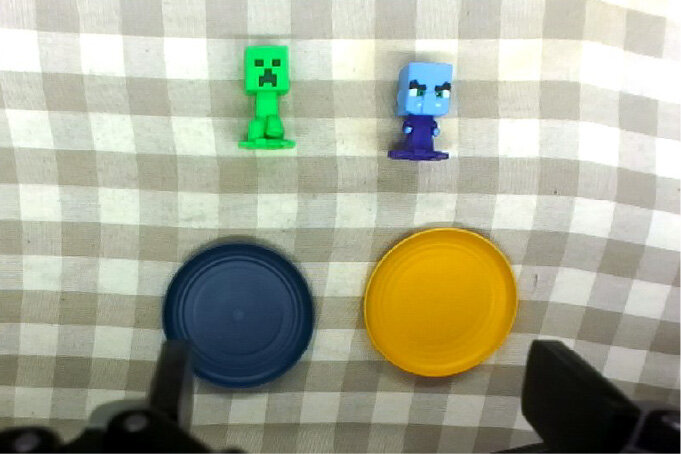} & Creeper $\to$ plate\idpromptsep ``pick up the creeper and place onto the purple plate'' & +50\% pick creeper toy, +50\% place onto purple plate\\
\includegraphics[width=\scenewidthland]{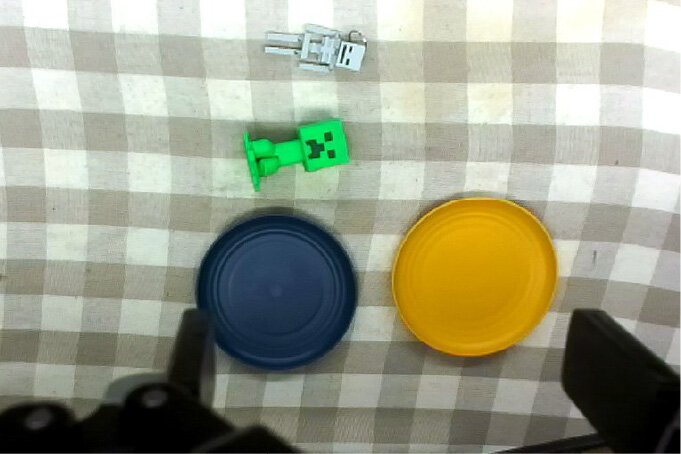} & Largest toy $\to$ plate\idpromptsep ``pick up the largest toy and place onto the purple plate'' &+50\% pick creeper, +50\% place onto purple plate, $-$20\% if attempt to place on distractor & \includegraphics[width=\scenewidthland]{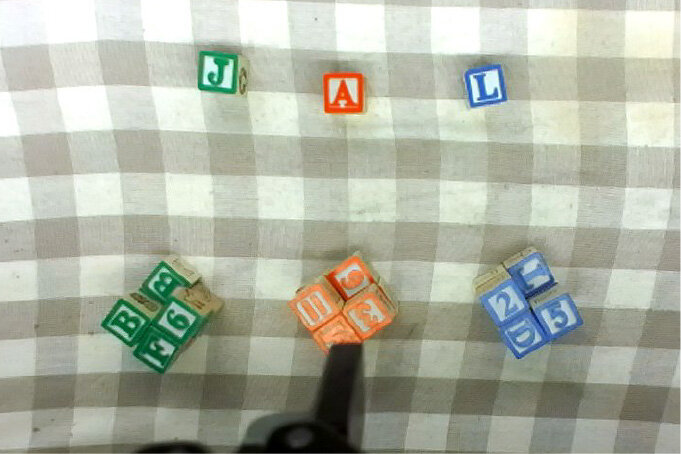} & Red A $\to$ color pile\idpromptsep ``pick up the red A and place on same color pile'' & +50\% pick red A block, +50\% place onto red pile, $-$20\% knock pile over\\
\includegraphics[width=\scenewidthland]{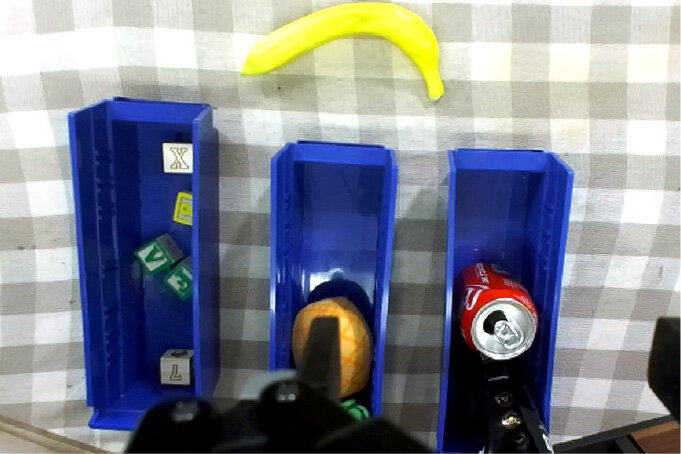} & Banana $\to$ box\idpromptsep ``pick up the banana and put it in the box'' & +50\% place banana into any box, +50\% place into box with fruit (aims to test common sense of human selection) & \includegraphics[width=\scenewidthland]{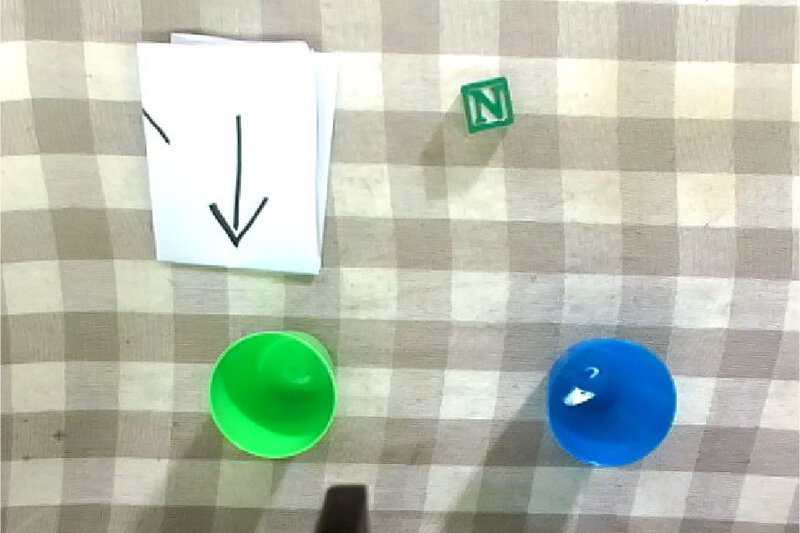} & N block $\to$ indicated cup\idpromptsep ``put the N block into the cup pointed to by the arrow'' & +50\% grasp N block, +50\% place into cup pointed at\\
\includegraphics[width=\scenewidthland]{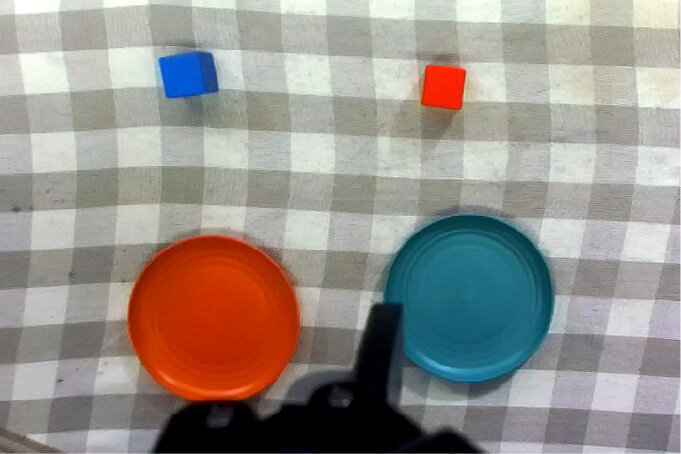} & Sort blocks by color\idpromptsep ``sort the blocks into opposite color plates'' & +10\% per block touched, +40\% per correct place & \includegraphics[width=\scenewidthland]{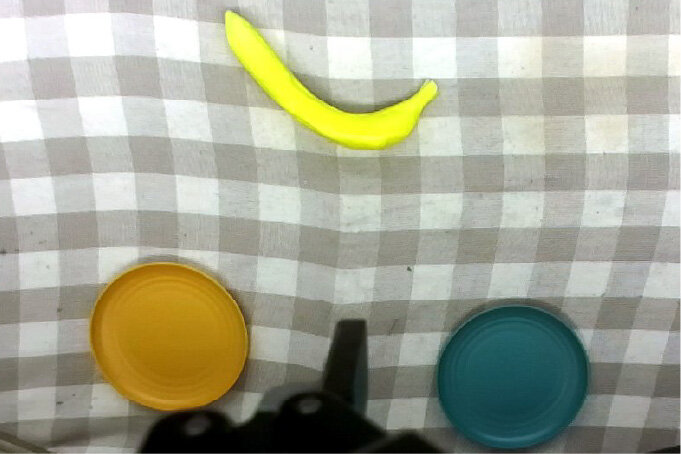} & Banana $\to$ matching plate\idpromptsep ``place banana into plate has similar color'' & +50\% pick banana, +50\% place into orange plate \\
\midrule
\multicolumn{6}{@{}l}{\emph{Multi-step}} \\
\includegraphics[width=\scenewidthland]{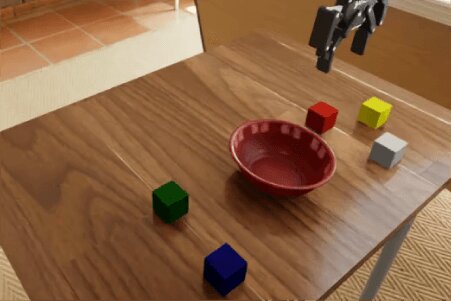} & Color cubes $\to$ bowl (sim)\idpromptsep ``put 3 cubes into the bowl'' & For up to 3 cubes (normalized to 100\%): +5\% approach cube, +10\% grasp, +10\% approach bowl with cube, +15\% place & \includegraphics[width=\scenewidthland]{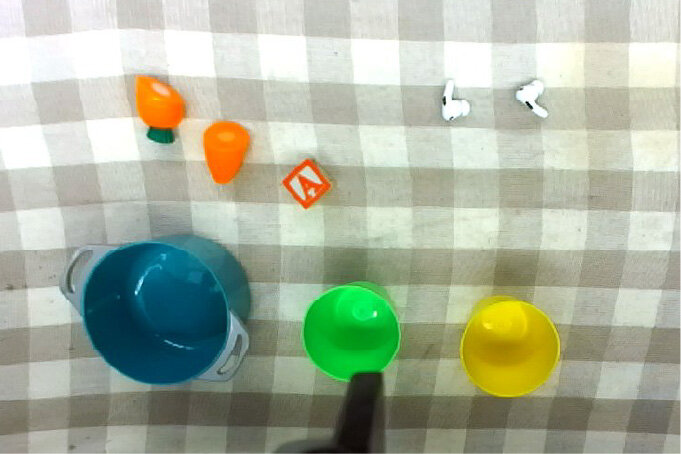} & AirPods $\to$ cup\idpromptsep ``place airpods into the yellow cup'' & +25\% per AirPods picked, +25\% per place, $-$20\% distractor \\
\includegraphics[width=\scenewidthland]{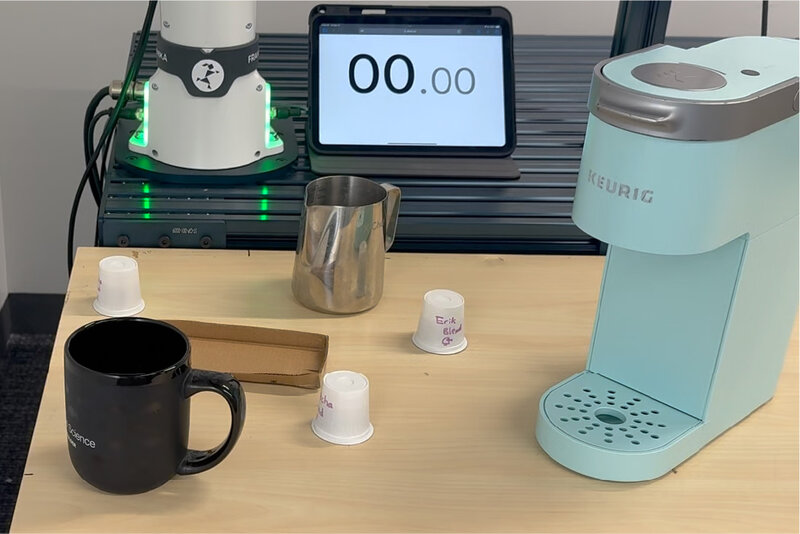} & Pack pods $\to$ tray$^\dagger$\idpromptsep ``pack the coffee pods onto the rectangular tray'' & For each of the 3 pods: +3.33\% approach, +15\% grasp, +0\% place not in tray, +15\% place touching tray & \includegraphics[width=\scenewidthland]{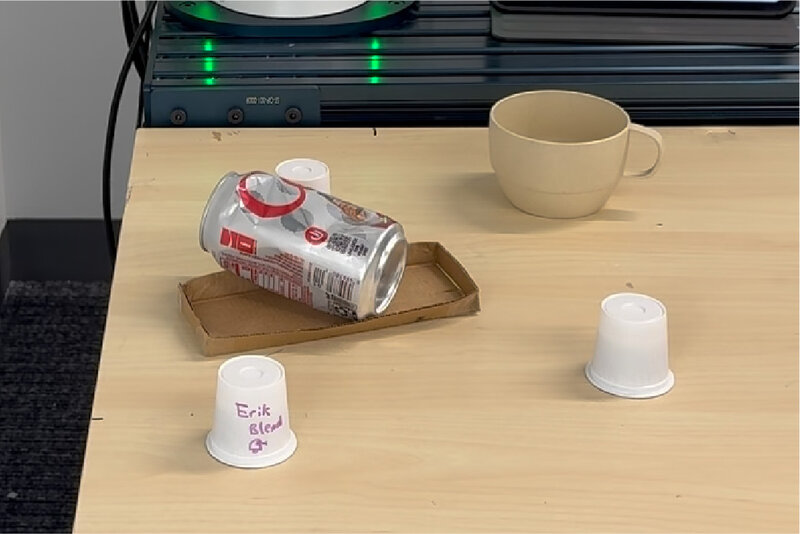} & Pack pods $\to$ tray (obs.)$^\dagger$\idpromptsep ``pack the coffee pods onto the rectangular tray'' & +12.5\% pick can, +12.5\% place s.t. it doesn't obstruct tray (or +25\% for clearing can obstruction without pick/place), for each of 3 pods: +5\% for approaching pod, +10\% for correct pick, +10\% for correct place into tray \\
\includegraphics[width=\scenewidthland]{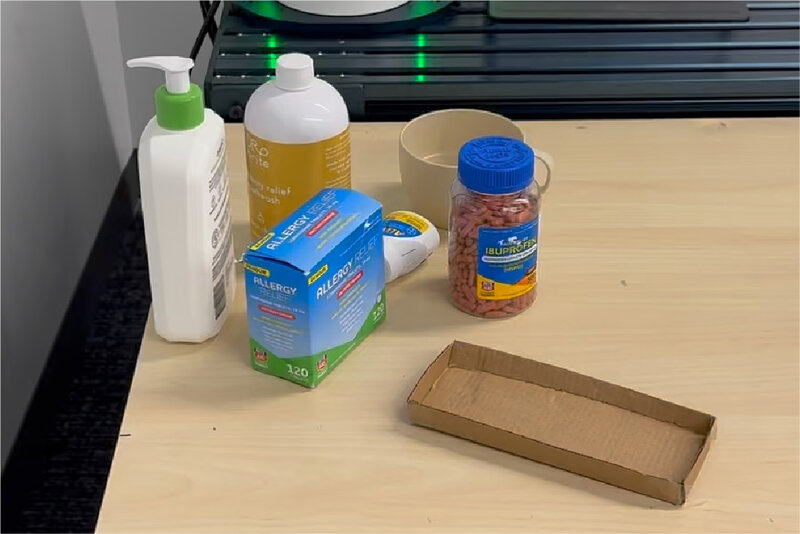} & Aleve bottle $\to$ tray (obs.)$^\dagger$\idpromptsep ``put the small white aleve bottle into the cardboard tray'' & +10\% pick an obstacle object, +10\% place obstacle s.t. unobstructs aleve, +30\% pick aleve bottle (+50\% if picked without clearing obstacles), +50\% place bottle in tray & \includegraphics[width=\scenewidthland]{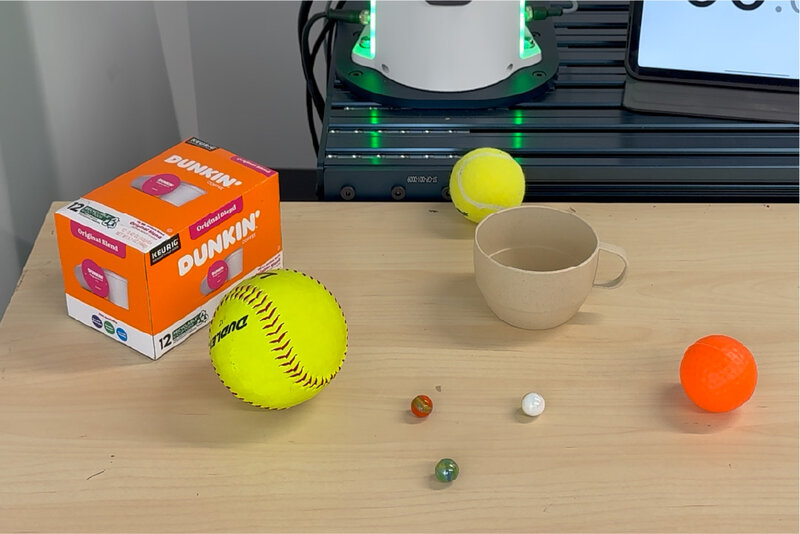} & Three marbles $\to$ cup$^\dagger$\idpromptsep ``put only the marbles in the cup'' & +16.67\% for each pick of a marble, +16.67\% for each place of a marble into the cup
 \\
\includegraphics[width=\scenewidthland]{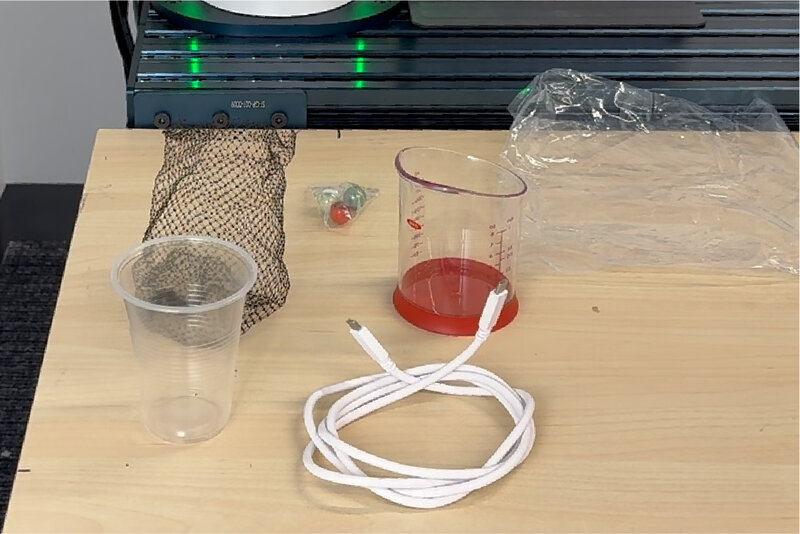} & Marbles + cable$^\dagger$\idpromptsep ``put the small plastic bag of marbles into the black mesh bag, and the cable on top of the empty large plastic bag'' & wire: +5\% approach, +20\% stable pick, +25\% stable place atop plastic; marbles pouch: +5\% approach, +20\% pick, +25\% place into mesh bag & & & 
\end{longtable}

\subsection{VLM Prompting Details}
\label{appendix:vlm-prompting}
As part of the perception module in \S\ref{sec:perception}, we use the following prompt for object detection and goal grounding:

\begin{lstlisting}
Perform two tasks on this image based on the task instruction: "{task_instruction}".

TASK 1 - OBJECT DETECTION:
Detect and return bounding boxes for objects in the image.
- DO NOT include the robot, robot gripper, or table surface
- DO NOT include objects or surfaces irrelevant to the task or too far away to matter (e.g. walls, things on the wall that are far away, things on the floor below the table, people who might be in the scene, etc.)
- Limit to 25 objects
- If an object appears multiple times, name them by unique characteristics (color, size, position, etc.). If they seem the same, then just use numbers (e.g. 'soda_can1' and 'soda_can2', ... for identical-looking soda cans)
- An object **cannot have the same name** as another under any circumstance.
- Format: normalized coordinates 0-1000 as integers

Be very careful to identify objects and name them in a way that's relevant to the task. If the task involves picking up a red apple, make sure that 'red' appears in the name of the apple.


TASK 2 - TASK TRANSLATION:
Translate this natural language instruction into some conjunction of formal predicates:

AVAILABLE PREDICATES:
- on(movable, surface): Object A is placed on top of object B

It is very important that the goal is exact: use your visual recognition, common-sense and reasoning abilities to make sure the goal expression is perfectly accurate.

For instance - for the task "throw away the trash in the bin" when there is a bin, an open empty chips packet, an empty soda can, a closed and full soda bottle, and several full candy bars on the table, the goal should be:

"predicates": [
    {{"name": "on", "args": ["chips_packet", "bin"]}},
    {{"name": "on", "args": ["soda_can", "bin"]}},
]

This is because only the chips and soda are empty and clearly trash. Everything else is still usable!


Return a single JSON object with this structure (no code fencing):
{{
    "bboxes": [
        {{"box_2d": [ymin, xmin, ymax, xmax], "label": "object name"}},
        ...
    ],
    "predicates": [
        {{"name": "predicate_name", "args": ["object1", "object2"]}},
        ...
    ]
}}

Use the object labels you detect in Task 1 when creating predicates in Task 2.
Only reference objects that you actually detected in the image.

\end{lstlisting}

\end{landscape}
\normalsize
\twocolumn

\end{document}